\documentclass{article}

\usepackage[preprint]{neurips_2026}

\usepackage[utf8]{inputenc}
\usepackage[T1]{fontenc}
\usepackage{hyperref}
\usepackage{url}
\usepackage{booktabs}
\usepackage{amsfonts}
\usepackage{nicefrac}
\usepackage{microtype}
\usepackage{xcolor}
\usepackage{amsmath}
\usepackage{amssymb}
\usepackage{graphicx}
\usepackage{subcaption}
\usepackage{algorithm}
\usepackage{algorithmic}


\title{Multi-Feature Riemannian Hypergraph for Online Test-Time Adaptation of Motor Imagery Brain-Computer Interface}

\author{%
  Siqi Li$^{1,2}$ \quad
  Zhi Li$^{3}$ \quad  
  Tong Liu$^{3}$ \quad
  Shuai Zhang$^{3}$ \quad
  Yanfei Jia$^{4}$ \\
  \AND
  Zhiqiang Yi$^{4}$ \quad
  Jue Xie$^{3}$ \quad
  Ni Ji$^{5,2}$\thanks{Corresponding author: niji@cibr.ac.cn} \\
  \\
  \\
  $^1$Peking University \\
  $^2$Chinese Institute for Brain Research, Beijing \\
  $^3$NeuCyber Neurotech \\
  $^4$Beijing Medical University \\
  $^5$Chinese Academy of Medical Sciences \& Peking Union Medical College \\
}

\begin{document}

\maketitle

\begin{abstract}
In clinical motor imagery brain-computer interface (MI-BCI) decoding, cross-day transferability and online operation remain two critical challenges. Hypergraphs can improve transferability by capturing higher-order sample relationships, yet existing hypergraph-based methods for online emotion recognition neglect the cross-day benefits of Riemannian geometry widely adopted in EEG transfer learning. To bridge this gap, we propose the Multi-feature Riemannian Hypergraph (MRieHy), a framework tailored for online test-time adaptation in MI-BCI decoding that leverages Riemannian geometry to strengthen cross-day transferability. MRieHy first computes Riemannian means of covariance matrices from cross-day training data to align multi-day distributions. It then constructs a hypergraph over covariance matrices using Riemannian distance, complemented by a second hypergraph over deep features built with cosine similarity. The two hypergraphs are fused via adaptively learned combination weights, jointly optimized with the label projection matrices. During online testing, MRieHy maintains a first-in-first-out buffer of recent samples, performs Riemannian alignment on the buffered data, and decodes with the learned hypergraph. Extensive experiments on a private four-class ECoG dataset and two public four-class EEG datasets validate that MRieHy achieves notable performance gains over state-of-the-art baselines.
\end{abstract}

\section{Introduction}\label{sec:introcudtion}

Motor imagery brain-computer interfaces (MI-BCIs) represent a transformative neurotechnology that enables direct communication between the human brain and external devices through the decoding of imagined movement patterns without physical execution \cite{Padfield2019EEGBasedBI}. For individuals with spinal cord injuries, amyotrophic lateral sclerosis, or stroke-induced paralysis, MI-BCIs offer a vital channel for environmental interaction, communication, and neurorehabilitation \cite{Elashmawi2024ACR,Chen2025BraincomputerII,Liu2025RecentAO}.

Despite remarkable progress, the practical deployment of MI-BCIs faces a critical challenge: the inherent non-stationarity of neural signals across time. Brain signals fluctuate due to factors like varying attention and fatigue levels \cite{Shenoy2006TowardsAC}, typically necessitating a supervised calibration session before each use. This recurring requirement creates significant barriers for both patients with limited cognitive resources and the general public, who expect plug-and-play usability \cite{Zanini2018TransferLA}.

This challenge is compounded by real‑time online operation, where signals stream continuously and demand immediate feedback. Prior work shows that even with calibration, many users still experience substantial performance degradation during online control, caused by the domain shift from offline calibration to real‑time conditions \cite{Vidaurre2011MachineLearningBasedCC}. Consequently, methods enabling effective domain adaptation from unlabeled streaming test‑time data are essential for practical deployment.

Hypergraphs extend traditional graphs by modeling higher-order relationships among multiple entities simultaneously, going beyond simple pairwise connections \cite{Zhou2006LearningWH}. By capturing such overall higher-order sample relationships, hypergraphs can improve generalization relative to simple feature extraction methods \cite{Ye2022OnlineEE}. Recent studies demonstrate their effectiveness in online transfer learning for emotion recognition using multimodal physiological signals such as EEG \cite{Ye2022OnlineEE,Pan2023MultimodalPS,Pan2024OnlineMF}.

However, these methods do not exploit the cross-day transfer benefits of Riemannian geometry, a common technique in EEG transfer learning. They also assume labeled data are available for online adaptation, which is rarely true in MI‑BCI applications. Moreover, the drivers of cross‑day variability may differ from those in emotion recognition, and MI‑BCI systems demand responses on a much shorter timescale.

We propose a novel framework, Multi-feature Riemannian Hypergraph (MRieHy), for online test-time adaptation in MI-BCI decoding. MRieHy constructs hypergraphs from deep neural features and covariance matrices of EEG/ECoG signals, then fuses them for the final prediction. Inspired by Riemannian geometry for transfer learning, it adopts a Riemannian metric-based similarity for hypergraph construction and applies Riemannian alignment to align multi-day training data with streaming test data, directly mitigating cross-day distribution shifts. Our contributions include:

$\bullet$ We propose a Riemannian metric-based similarity for hypergraph construction, overcoming the limitations of conventional cosine-similarity-based hypergraph frameworks.

$\bullet$ MRieHy’s multi-feature architecture combines the complementary strengths of covariance representations, which excel with limited training data, and deep features, which become advantageous with larger datasets.

$\bullet$ To the best of our knowledge, this is the first work to validate online test-time adaptation on human MI-BCI ECoG data. We demonstrate that MRieHy outperforms state-of-the-art baselines on ECoG and EEG benchmarks, as confirmed by detailed ablation analyses.

\section{Related Work}
\subsection{Riemannian Geometry for BCI}
As the covariance matrices of spatiotemporal EEG signals are symmetric positive definite (SPD), they lie on the SPD manifold. On this manifold, Riemannian metrics appropriately define distances and mean functions, endowing BCIs with key advantages: equivalence between sensor and source spaces, robustness of the geometric mean to artifacts, and strong cross-day and cross-subject generalization capabilities \cite{Congedo2017RiemannianGF}.

Taking advantages of the Riemannian geometry, Barachant et al.\ \cite{Barachant2012MulticlassBI} implement the minimum distance to mean (MDM) classification algorithm with Riemannian distance and Riemannian mean, as well as the linear discriminant analysis in Riemannian tangent space. Zanini et al.\ \cite{Zanini2018TransferLA} improve the MDM algorithm by considering the dispersion of each class in addition to the mean of each class, and apply their algorithm to cross-day and cross-subject BCI classification. Riemannian geometry is further combined with graph to preserve the Riemannian locality of data structure for BCI transfer learning \cite{Xu2024RiemannianLP}, and is joint with Long Short-Term Memory network for inter-day three-dimensional hand trajectories decoding from ECoG signals \cite{Eyvazpour2025AGM}. Relative to convolutional neural networks, Riemannian decoding methods perform comparably in within-day, cross-day, and cross-subject settings \cite{Eder2024BenchmarkingBI}.

\subsection{Online Test-Time Adaptation}
Domain gaps frequently exist between training and testing data. At the same time, in real-world scenarios, obtaining calibration data from the same domain as the testing data is often difficult, and testing data typically arrives in an online stream. To overcome this challenge, online test-time adaptation methods have been proposed. These methods utilize only unlabeled testing data for adaptation during online inference \cite{Liang2023ACS}. This characteristic makes online test-time adaptation particularly suitable for Brain-Computer Interfaces (BCIs), as a primary challenge in real-world BCI deployment is overcoming day-to-day instability without calibration \cite{Zanini2018TransferLA}.

Specifically, Riemannian alignment or Euclidean alignment of the covariance matrices of the EEG signals is used to increase the data similarity across days and subjects. In batch normalization in deep learning, the statistics are estimated and updated using the test-time data, instead of inheriting from training statistics directly \cite{Bougrain2021GuidelinesTU,Wimpff2023CalibrationfreeOT,Ouahidi2024UnsupervisedAD,Wimpff2025FineTuningSF}. To update the model during testing, entropy minimization is used to increase the confidence of the model's prediction. Meanwhile, marginal distribution regularization or source model guidance can be applied to prevent the model from assigning all test samples to a single class and from becoming overly confident in incorrect predictions, which could be the results of entropy minimization \cite{Wimpff2023CalibrationfreeOT,Li2023TTIMETI,Peng2025TestTimeAF}.

\subsection{Hypergraph Learning}
As a generalization of graph, hypergraph can represent complex relationship between objects other than pairwise relationship \cite{Zhou2006LearningWH}. Excelling at multi-modal learning, hypergraph-based algorithms are applied to three-dimensional object classification with multiple view data \cite{Zhang2018InductiveML}, as well as emotional recognition with both personality and physiological signals \cite{Zhao2018PersonalityAwarePE}.

Yalan et al.\ \cite{Ye2022OnlineEE} adapt hypergraph-based algorithms for online cross-subject transfer learning in emotion recognition using electrocardiogram (ECG) signals by adding hyperedges that connect incoming samples from unknown subjects to a hypergraph constructed from data of known subjects. Furthermore, the multimodal learning capability of hypergraph-based methods has been exploited to fuse EEG, ECG, and galvanic skin response signals for online cross-subject emotion recognition \cite{Pan2023MultimodalPS,Pan2024OnlineMF}. However, these methods require labeled data to update the hypergraph when performing online transfer learning for previously unseen subjects. The labeled test-time data may not be available in real-world BCI applications. Furthermore, their hypergraphs are constructed using k-nearest neighbors with cosine similarity, and thus do not leverage the advantages of Riemannian geometry in cross-day and cross-subject settings.

\begin{figure}[ht]
    \centering
    \includegraphics[width=0.95\linewidth]{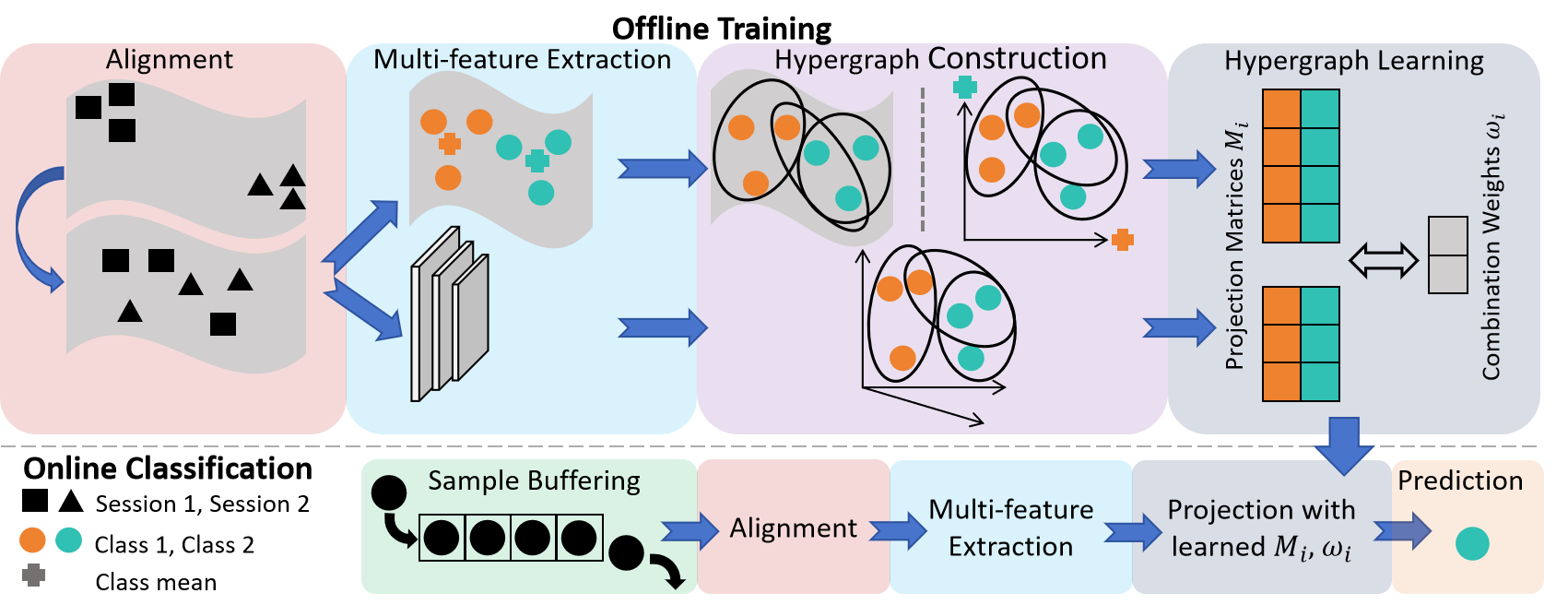}
    \caption{Schematic illustration of the Multi-feature Riemannian Hypergraph (MRieHy) framework.} %In the offline training phase, the framework aligns multi-day data and extracts dual feature sets---covariance matrices and deep representations---to construct hypergraphs using Riemannian and cosine metrics. It subsequently optimizes projection matrices and adaptive combination weights for graph fusion. During online classification, unlabeled test samples are buffered, aligned, and projected via the pre-trained parameters to facilitate efficient label prediction.}
    \label{fig:MRieHy}
\end{figure}
\section{Method}\label{sec:method}
The Multi-feature Riemannian Hypergraph (MRieHy) framework, illustrated in Figure \ref{fig:MRieHy}, enables robust online test-time adaptation for MI-BCI decoding. During offline training, it aligns multi-day data using Riemannian means and extracts both covariance matrices and deep features to construct two hypergraphs---utilizing Riemannian distance-based similarity for covariance matrices and cosine similarity for deep features. It then learns projection matrices and adaptive combination weights to fuse these graphs. For online classification, incoming unlabeled samples are buffered and aligned to the training distribution before features are extracted and projected using the pre-learned matrices and weights to predict labels.

\subsection{Preliminary}
In the $N$-class MI-BCI classification task, there is training data from $K$ labeled source days $\{D_\mathcal{S}^k\}_{k=1}^K$ of a certain subject. In the $k$-th source day, the $i$-th sample $\mathbf{X}_i^k \in \mathbb{R}^{C \times W}$ is obtained via channel-wise centering after sliding-window processing of a trial $\mathbf{L}_j^k \in \mathbb{R}^{C \times T}$, where $C$ is the channel number, $T$ is the time samples of the trial, $W$ is window length, and $W\leq T$. It corresponds to label $y_i^k \in \{1, 2, \cdots, N\}$. Data from all source days is aligned and used to train hypergraph-based classifier $f_\mathcal{S}$. When testing, a sequence of unlabeled samples $\{\mathbf{X}_1^\mathcal{T}, \mathbf{X}_2^\mathcal{T}, \cdots\}$ from target day $D_\mathcal{T}$ arrive in an online manner, and are dynamically stored in a first-in-first-out (FIFO) buffer with size $b$. Online test-time adaptation is then utilized to leverage the labeled knowledge implied in $f_\mathcal{S}$ to infer the labels of these samples in $D_\mathcal{T}$ under distribution shift \cite{Liang2023ACS}.

\subsection{Riemannian Alignment}
Favored for its simplicity and effectiveness, covariance alignment functions as a standard transfer learning approach in BCI decoding \cite{He2018TransferLF,Zanini2018TransferLA,Kostas2020ThinkerIE}. The method assumes that differences between days can be eliminated by a whitening process with the mean of covariance matrices \cite{Tomioka2006AdaptingSF,Reuderink2011ASB,Zanini2018TransferLA}. Specifically, the mean $\mathbf{\overline{R}}^k$ of the covariance of the samples in day $k$ in the sense of Euclidean (equation \eqref{eq:cov_matrix_mean} left) or Riemannian geometry (equation \eqref{eq:cov_matrix_mean} right) is first calculated:

\begin{equation}
\mathbf{\overline{R}}^k = \frac{1}{l} \frac{1}{W-1}\sum_{i=1}^l \mathbf{X}_i^k \cdot \mathbf{X}_i^{k\top}, \hspace{0.5cm}\mathbf{\overline{R}}^k = \arg\min_R\sum_{i=1}^ld_g^2(R, \frac{1}{W-1}\mathbf{X}_i^k \cdot \mathbf{X}_i^{k\top})
\label{eq:cov_matrix_mean}
\end{equation}

% \begin{equation}
% \mathbf{\overline{R}}^k = \arg\min_R\sum_{i=1}^ld_g^2(R, \frac{1}{W-1}\mathbf{X}_i^k \cdot \mathbf{X}_i^{k\top})
% \label{eq:cov_matrix_mean_Rie}
% \end{equation}

where $l$ is the sample number in day $k$, and $d_g$ represents Riemannian distance. Then, each trial is aligned by:
\begin{equation}
\widetilde{\mathbf{X}}_i^k = (\mathbf{\overline{R}}^{k})^{-\frac{1}{2}} \cdot \mathbf{X}_i^k
\label{eq:trial_align}
\end{equation}

The aligned sample $\widetilde{\mathbf{X}}_i^k \in \mathbb{R}^{C \times W}$ is used for the following decoding processes.

\subsection{Riemannian Hypergraph Construction}

For simplicity, the aligned sample $\widetilde{\mathbf{X}}_i^k$ is written as $\mathbf{X}$ here. Using $\mathbf{X}$, multiple features are extracted for hypergraph learning. The covariance matrix $\mathbf{F}_i^{co} \in \mathbb{R}^{C \times C}$ is computed as
\begin{equation}
\mathbf{F}_i^{co} = \frac{1}{W-1}\sum_{t=1}^W (\mathbf{X}_{\cdot,t} - \overline{\mathbf{X}})(\mathbf{X}_{\cdot,t} - \overline{\mathbf{X}})^\top
\label{eq:eeg_cov_np}
\end{equation}

where $\overline{\mathbf{X}} = \frac{1}{W} \sum_{t=1}^W \mathbf{X}_{\cdot,t} \in \mathbb{R}^C$ denotes the temporal mean of the channel signals. $\mathbf{F}_i^{co}$ can be seen as a point on the Riemannian manifold of symmetric positive definite (SPD) matrices \cite{Zanini2018TransferLA}.
We use $\mathbf{f}_i^{co} \in \mathbb{R}^{C\cdot C}$  to represent the vectorized covariance matrix $\mathbf{F}_i^{co}$. Meanwhile, the aligned samples and the corresponding labels are used to train BaseNet, a lightweight yet powerful model for EEG decoding \cite{Wimpff2023CalibrationfreeOT,Wimpff2025FineTuningSF}. The deep feature $\mathbf{f}_i^{deep} \in \mathbb{R}^d$ before the final linear layer of BaseNet is extracted as another feature, where $d$ represents the dimension of it.

Each of these two features serves as the basis for a hypergraph. To construct the hypergraph, we employ two distinct classes of methods to measure the similarity $s(v_i, v_j)$ between hypergraph vertices $v_i$ and $v_j$ corresponding to samples $\mathbf{X}_i$ and $\mathbf{X}_j$.

One class is based on the relationship between sample features directly. Belonging to this class is the cosine similarity (Cos), which can be applied to both covariance matrix $\mathbf{f}_i^{co}$ and deep feature $\mathbf{f}_i^{deep}$:
\begin{equation}
s(v_i, v_j)=\frac{\langle\mathbf{f}_i,\mathbf{f}_j\rangle}{|\mathbf{f}_i|\cdot|\mathbf{f}_j|}
\label{eq:cos_sim}
\end{equation}

Two Riemannian versions to measure the relationship between sample features directly can be applied to the covariance matrix $\mathbf{F}_i^{co}$. When $\mathbf{F}_i^{co}$ is projected to the tangent space at the Riemannian center of all covariance matrices, the tangent space cosine similarity (TanCos) is computed with
\begin{equation}
s(v_i, v_j)=\frac{\langle\text{tan}(\mathbf{F}_i^{co}),\text{tan}(\mathbf{F}_j^{co})\rangle}{|\text{tan}(\mathbf{F}_i^{co})|\cdot|\text{tan}(\mathbf{F}_j^{co})|}
\label{eq:tan_cos_sim}
\end{equation}

Another version is to use the Gaussian kernel similarity based on Riemannian distance $d_g$ without projecting to tangent space (GauRie):
\begin{equation}
s(v_i, v_j)=\exp\left(-\frac{d_g^2(\mathbf{F}_i^{co},\mathbf{F}_j^{co})}{2\sigma^2}\right),\hspace{0.3cm} \sigma = \frac{1}{M^2} \sum_{i=1}^M \sum_{j=1}^M d_g(\mathbf{F}_i^{co},\mathbf{F}_j^{co})
\label{eq:SPD_cos_sim}
\end{equation}

where $M$ is the number of vertex pairs, and $\sigma$ equals to the mean distance.

The second class is inspired from the MDM algorithm. We calculate the Euclidean means $\overline{\mathbf{R}}_1^{Eu},\cdots,\overline{\mathbf{R}}_N^{Eu}$ or the Riemannian means $\overline{\mathbf{R}}_1^{Rie},\cdots,\overline{\mathbf{R}}_N^{Rie}$ of the covariance matrices with labels $1,\cdots, N$, using equations with identical form as \eqref{eq:cov_matrix_mean}. Then, for each covariance matrix  $\mathbf{F}_i^{co}$,  we compute the Euclidean distances to all Euclidean means (EuDM), the Riemannian distances to all Riemannian means (RieDM), or the Euclidean distances to all Riemannian means after projected to the tangent space at the mean of all covariance matrices (TanDM), forming $N$-dimension distance vector $\mathbf{d}_i^{Eu}\in\mathbb{R}^N$,  $\mathbf{d}_i^{Rie}\in\mathbb{R}^N$, or $\mathbf{d}_i^{tan}\in\mathbb{R}^N$. Finally, the similarity $s(v_i,v_j)$ between vertices $v_i$ and $v_j$ is calculated using the cosine similarity of each type of the distance to means $\mathbf{d}_i$ and $\mathbf{d}_j$.

Using the pairwise similarities from deep features and covariance matrix, we model the relationships among hypergraph vertices with two hypergraphs, $\mathcal{G}_{deep}=(\mathcal{V}_{deep},\mathcal{E}_{deep},\mathbf{W}_{deep})$, and $\mathcal{G}_{co}=(\mathcal{V}_{co},\mathcal{E}_{co},\mathbf{W}_{co})$, where $\mathcal{V}$ represents the vertex set, $\mathcal{E}$ represents the hyperedge set, and $\mathbf{W}$ represents the diagonal matrix of hyperedge weight. We call $\mathcal{G}_{co}$ \textit{Riemannian hypergraph} when it is built with similarities calculated with equation \eqref{eq:tan_cos_sim} or equation \eqref{eq:SPD_cos_sim}, or calculated from $\mathbf{d}_i^{Rie}$ or $\mathbf{d}_i^{tan}$, since they are Riemannian-based. In each hypergraph, we create a hyperedge for each vertex to connect it to its $k$ nearest neighbors based on the similarities. This hyperedge captures the relationship among the samples it connects. Given the constructed hypergraph $\mathcal{G}$, the incidence matrix $\mathbf{H}$ is obtained by computing each entry as:
\begin{equation}
    \mathbf{H}(v,e)=\left\{
    \begin{aligned}
        1,\hspace{0.5cm} v\in e\\
        0, \hspace{0.5cm} v\notin e
    \end{aligned}
    \right.
\end{equation}

In this framework, for any vertex $v \in \mathcal{V}$ and hyperedge $e \in \mathcal{E}$, their degrees are defined respectively by
\begin{equation}
d(v) = \sum_{e \in \mathcal{E}} \mathbf{W}(e)\mathbf{H}(v, e),\hspace{0.5cm}\delta(e) = \sum_{v \in \mathcal{V}} \mathbf{H}(v, e)
\label{eq:vertex_degree}
\end{equation}

Using these quantities, we construct two diagonal matrices, $\mathbf{D}^v$ and $\mathbf{D}^e$, containing the vertex degrees and hyperedge degrees, respectively. The hyperedge weights are initialized uniformly as $1/n_e$, where $n_e$ denotes the total number of hyperedges.

\subsection{Multi-feature Hypergraph Learning}\label{subsec:mf_hg_learn}
Let $\mathbf{Z}=[\mathbf{f}_1,\mathbf{f}_2,\cdots,\mathbf{f}_n]$ be the concatenation of all sample features, where $n$ is the total sample number for training. Following \cite{Zhang2018InductiveML}, the goal of hypergraph learning is to learn a regularized matrix $\mathbf{M}$ to project the feature matrix $\mathbf{Z}$ to the label space to discriminate different categories.

To learn the projection matrix $\mathbf{M}$ for a single hypergraph, a cost function $\mathbf{\Psi}$ is introduced, aggregating the hypergraph Laplacian regularizer $\mathbf{\Omega}(\mathbf{M})$, the empirical loss $\mathcal{R}_{emp}(M)$, and a regularization term on $\mathbf{M}$ given by $\mathbf{\Phi}(\mathbf{M})$:
\begin{equation}
    \mathbf{\Psi}=\{\mathbf{\Omega}(\mathbf{M})+\lambda\mathcal{R}_{emp}(M)+\mu\mathbf{\Phi}(\mathbf{M})\}
\label{eq:cost_func}
\end{equation}

In the context of multi-feature hypergraphs, the cost function $\overline{\Psi}$ for learning the projection matrices $\mathbf{M}_h$ from all hypergraphs comprises two components: the aggregated learning costs across individual hypergraphs and a regularization term $\boldsymbol{\Gamma}$ applied to the combination weights $\boldsymbol{\omega}$:
\begin{equation}
\overline{\Psi} = \sum_{h=1}^{m} \omega_h \left\{ \Omega(\mathbf{M}_h) + \lambda \mathcal{R}_{emp}(\mathbf{M}_h) + \mu \Phi(\mathbf{M}_h) \right\} + \eta \Gamma(\boldsymbol{\omega})
\end{equation}

where $m$ denotes the number of hypergraphs (equaling 2 in this research), and $h$ indexes the specific hypergraph. The projection matrices $\mathbf{M}_h$ for all hypergraphs are learned by optimizing this cost function. The detailed process is described in the appendix.

\subsection{Online Test-Time Adaptation}

We deploy our multi-feature Riemannian hypergraph in an online setting where test samples arrive one by one and are stored in a FIFO buffer of size $b$. For each new sample, we align the buffer contents using Riemannian or Euclidean alignment (starting from the very first sample) for domain adaptation, and extract its covariance matrix $\mathbf{f}_t^{co}$ and deep feature $\mathbf{f}_t^{deep}$. The final prediction is then obtained via
\begin{equation}
\hat{\mathbf{y}}_t=\omega_{deep}(\mathbf{f}_t^{deep})^\top\mathbf{M}_{deep}+\omega_{co}(\mathbf{f}_t^{co})^\top\mathbf{M}_{co}\in\mathbb{R}^N,
\end{equation}
\begin{equation}
\hat{y}_t=\arg\max_c\hat{\mathbf{y}}_t(c)
\end{equation}
$\mathbf{M}_1,\cdots,\mathbf{M}_m$ in section \ref{subsec:mf_hg_learn} is specified as $\mathbf{M}_{deep}$ and $\mathbf{M}_{co}$ here, as well as $\omega_1,\cdots,\omega_m$.

\section{Experiments}\label{sec:experiments}
\subsection{Dataset}

We apply our method to three MI-BCI datasets with two modalities: a private ECoG dataset ECoG128, and two public EEG datasets BCI Competition IV 2a \cite{katb-zv89-24} and Stieger2021 \cite{Stieger2021ContinuousSR}.

The ECoG128 dataset was acquired using a flexible electrode array of 128 channels, which was implanted epidurally over the primary motor and somatosensory cortices in the left hemisphere of a participant with tetraplegia due to spinal cord injury, covering an area of $43\times43\text{mm}^2$.
The participant performed MI-BCI tasks with 4 classes. Data were collected across eight days. The number of trials per day ranged from 200 to 520, yielding a total of 2,840 trials. The neural signals (600 Hz sampling rate) are bandpass-filtered between 4 Hz and 100 Hz and subsequently resampled to 256 Hz, followed by a sliding window process with 1s window length and 1s stride. The samples after window sliding are then channel-wise centered, and are shuffled randomly.

BCI Competition IV 2a dataset \cite{katb-zv89-24} consists of EEG data recorded at 250 Hz using 22 electrodes from 9 subjects performing 4-class motor imagery task. It was recorded in two sessions on different days with 288 trials per session. The signals are low-pass filtered with a cutoff frequency of 40 Hz, and are channel-wise centered.

The Stieger2021 dataset \cite{Stieger2021ContinuousSR} comprises 1000 Hz EEG recordings covering two- and four-class motor imagery tasks. We randomly selected 7 subjects from the dataset, each with 11 sessions recorded on separate days and 450 trials per session. Since the two-class conditions are contained within the four-class ones, all trials were framed as a four-class task. We selected signals from 24 channels centered over the motor cortex, low-pass filtered them at 200 Hz, resampled to 250 Hz, segmented them with a 1 s sliding window and 1 s stride, and finally applied channel-wise centering and random shuffling.

\subsection{Experiment Setting}

For the ECoG128 dataset, models were trained on Days 3--5 recordings with an 80/20 random train/validation split. Online evaluation with cumulative accuracy was conducted independently on held-out Days 6--8 data. To study the effect of training duration (Section \ref{subsec:n_train_d}), separate models were trained on subsets spanning 1 to 5 days: Day 5, Days 4--5, Days 3--5, Days 2--5, and Days 1--5.

For the BCI Competition IV 2a dataset, subject‑specific models were trained on 80\% of Day 1, validated on the remaining 20\%, and tested online on all Day 2 recordings for cross‑day generalization.

The Stieger2021 dataset followed the same pipeline as ECoG128: Days 1--8 were used for training (80/20 split), and Days 9--11 were held out for online evaluation with cumulative accuracy. Likewise, training length was varied from 1 day (Day 8) up to 8 days (Days 1--8).

We compare MRieHy with BaseNet \cite{Wimpff2025FineTuningSF,Wimpff2023CalibrationfreeOT}, Riemannian minimum distance to mean (RieMDM) \cite{Kumar2024TransferLP}, and an ensemble BaseNet+RieMDM, where RieMDM class probabilities are derived via softmax of the negative Riemannian distances from the test covariance to each class mean.  
Additionally, we evaluate against test‑time adaptation methods including BN‑adapt \cite{Schneider2020ImprovingRA}, Tent \cite{Wang2021TentFT}, PL \cite{Lee2013PseudoLabelT}, CoTTA \cite{Wang2022ContinualTD}, and T‑TIME \cite{Li2023TTIMETI}. Euclidean cosine‑similarity hypergraph with Euclidean alignment (EuHy)g and its multi‑feature counterpart (MEuHy) are also included as hypergraph‑based baselines.

\begin{table*}[h]
    \centering
    \begin{subtable}{\textwidth}
        \centering
        \makebox[\textwidth][c]{
            \setlength{\tabcolsep}{2pt}
            \small  % 适当缩小字号，使表格更紧凑
            \begin{tabular}{l|cccc|cccccccc}
                \toprule
                \multicolumn{1}{c|}{Method} & \multicolumn{4}{c|}{ECoG128} & \multicolumn{7}{c}{Stieger2021 (EEG)} \\
                 & D6 & D7 & D8 & Avg & Subj1 & Subj12 & Subj26 & Subj28 & Subj30 & Subj46 & Subj50 & Avg \\
                \midrule
                BaseNet         & 63.8$\pm$0.8 & 58.5$\pm$1.2 & 65.5$\pm$1.7 & 62.6$\pm$0.9 & \underline{62.1$\pm$1.5} & 35.8$\pm$0.3 & 56.8$\pm$1.0 & 41.1$\pm$3.3 & \textbf{59.1$\pm$1.7} & 63.4$\pm$1.8 & 49.1$\pm$0.9 & 52.5$\pm$0.1\\
                RieMDM          & 62.0$\pm$0.4 & 56.4$\pm$1.1 & 64.4$\pm$0.4 & 60.9$\pm$0.6 & 48.4$\pm$0.3 & \underline{36.8$\pm$0.5} & 52.1$\pm$0.5 & 37.8$\pm$0.1 & 42.0$\pm$0.3 & 50.4$\pm$0.3 & 39.1$\pm$0.1 & 43.8$\pm$0.2\\
                BaseNet+RieMDM  & 63.8$\pm$0.9 & \underline{58.7$\pm$1.1} & \underline{65.7$\pm$1.5} & \underline{62.7$\pm$0.8} & \textbf{62.2$\pm$1.4} & 36.1$\pm$0.2 & 57.0$\pm$1.0 & 41.4$\pm$3.4 & \underline{58.9$\pm$1.7} & 63.5$\pm$1.8 & 49.2$\pm$0.8 & 52.6$\pm$0.1\\
                BN-adapt        & 62.5$\pm$0.1 & 56.9$\pm$1.2 & 64.3$\pm$1.1 & 61.2$\pm$0.3 & 54.3$\pm$0.3 & 34.9$\pm$1.1 & 55.9$\pm$0.4 & 42.1$\pm$1.4 & 52.8$\pm$2.1 & 56.6$\pm$1.0 & 45.5$\pm$1.2 & 48.9$\pm$0.7\\
                Tent            & 59.0$\pm$0.3 & 54.0$\pm$1.1 & 61.0$\pm$1.4 & 58.0$\pm$0.8 & 49.4$\pm$0.8 & 32.1$\pm$0.9 & 50.6$\pm$0.8 & 39.5$\pm$1.6 & 48.7$\pm$0.8 & 52.6$\pm$0.7 & 41.1$\pm$1.5 & 44.9$\pm$0.6 \\
                PL              & 63.0$\pm$0.7 & 57.9$\pm$1.0 & 64.0$\pm$1.4 & 61.6$\pm$0.5 & 55.2$\pm$0.3 & 35.5$\pm$1.2 & 56.8$\pm$0.4 & 42.7$\pm$1.4 & 53.6$\pm$1.8 & 57.1$\pm$0.9 & 45.8$\pm$1.1 & 49.5$\pm$0.7\\
                CoTTA           & 63.0$\pm$1.1 & 58.1$\pm$1.5 & 65.2$\pm$1.0 & 62.1$\pm$0.5 & 61.9$\pm$1.1 & 35.6$\pm$1.0 & 57.8$\pm$3.7 & 43.3$\pm$3.8 & 58.5$\pm$2.5 & \underline{64.5$\pm$1.6} & \textbf{50.6$\pm$1.0} & \underline{53.2$\pm$0.5}\\
                T-TIME          & \underline{63.8$\pm$0.7} & 58.3$\pm$0.5 & 64.5$\pm$1.1 & 62.2$\pm$0.3 & 55.6$\pm$0.4 & 35.7$\pm$1.1 & 57.0$\pm$0.2 & \underline{43.9$\pm$1.6} & 53.6$\pm$2.0 & 57.5$\pm$0.9 & 46.3$\pm$1.3 & 49.9$\pm$0.7 \\
                EuHy            & 49.2$\pm$5.6 & 44.6$\pm$5.7 & 43.4$\pm$5.6 & 45.7$\pm$5.6 & 26.6$\pm$1.1 & 23.7$\pm$0.4 & 36.7$\pm$1.2 & 16.9$\pm$0.0 & 25.7$\pm$1.9 & 27.8$\pm$1.7 & 29.9$\pm$0.4 & 26.8$\pm$0.7 \\
                MEuHy           & 60.6$\pm$2.3 & 54.1$\pm$1.7 & 54.9$\pm$3.2 & 56.5$\pm$2.3 & 59.9$\pm$1.5 & 34.4$\pm$0.5 & \underline{58.9$\pm$1.2} & 40.4$\pm$1.4 & 54.4$\pm$2.1 & 60.5$\pm$2.0 & 48.4$\pm$1.1 & 51.0$\pm$0.1 \\
                MRieHy          & \textbf{65.7$\pm$0.5} & \textbf{59.5$\pm$0.9} & \textbf{67.2$\pm$1.8} & \textbf{64.1$\pm$0.9} & 61.9$\pm$0.3 & \textbf{37.5$\pm$0.7} & \textbf{60.0$\pm$0.8} & \textbf{45.9$\pm$0.8} & 58.2$\pm$1.3 & \textbf{65.3$\pm$1.7} & \underline{49.6$\pm$0.3} & \textbf{54.1$\pm$0.4}\\
                \bottomrule
            \end{tabular}
        }
        \label{tab:ecog}
    \end{subtable}
    
    \vspace{0.3em}  % 上下表格之间的垂直间距
    
    \begin{subtable}{\textwidth}
        \centering
        \makebox[\textwidth][c]{
        \setlength{\tabcolsep}{2pt}        
        \small
        \begin{tabular}{l|*{10}{c}} % 10个居中列
            \toprule
            \multicolumn{1}{c|}{Method} & \multicolumn{10}{c}{BCI Competition IV 2a (EEG)}\\
            & Subj1 & Subj2 & Subj3 & Subj4 & Subj5 & Subj6 & Subj7 & Subj8 & Subj9 & Avg \\
            \midrule
            BaseNet         & \underline{74.0$\pm$4.9} & 58.0$\pm$2.7 & 82.9$\pm$3.3 & 52.3$\pm$9.3 & \underline{71.4$\pm$1.7} & 50.7$\pm$2.5 & 84.4$\pm$3.7 & \underline{70.8$\pm$1.8} & 69.1$\pm$4.8 & 68.2$\pm$1.4 \\
            RieMDM          & 63.4$\pm$1.3 & 48.4$\pm$1.1 & 68.6$\pm$1.6 & \underline{54.9$\pm$0.9} & 40.7$\pm$0.7 & 43.6$\pm$1.2 & 62.3$\pm$1.1 & 75.3$\pm$0.9 & 70.4$\pm$3.2 & 58.6$\pm$0.2 \\
            BaseNet+RieMDM  & 74.4$\pm$4.1 & \underline{58.3$\pm$3.0} & \textbf{83.1$\pm$3.7} & 52.3$\pm$9.1 & \textbf{71.6$\pm$1.6} & 51.6$\pm$2.5 & \underline{84.8$\pm$3.5} & 70.7$\pm$1.1 & \underline{70.8$\pm$4.2} & \underline{68.6$\pm$1.3} \\
            BN-adapt        & 69.7$\pm$2.8 & 57.3$\pm$0.9 & 76.9$\pm$7.9 & 46.5$\pm$8.1 & 70.3$\pm$3.6 & 51.7$\pm$3.3 & 79.7$\pm$11.4 & 64.8$\pm$3.0 & 64.8$\pm$1.6 & 64.6$\pm$2.8 \\
            Tent            & 63.2$\pm$1.9 & 51.7$\pm$2.4 & 71.6$\pm$5.1 & 45.1$\pm$6.9 & 65.3$\pm$1.6 & 50.0$\pm$0.3 & 74.9$\pm$8.8 & 59.5$\pm$0.7 & 60.0$\pm$2.1 & 60.1$\pm$0.9 \\
            PL              & 69.1$\pm$3.3 & 57.8$\pm$1.4 & 76.9$\pm$6.7 & 46.5$\pm$10.3 & 70.4$\pm$3.5 & 51.0$\pm$3.3 & 79.7$\pm$10.6 & 64.8$\pm$2.5 & 64.4$\pm$3.1 & 64.5$\pm$2.4 \\
            CoTTA           & 70.4$\pm$3.7 & 58.0$\pm$0.6 & 76.5$\pm$7.3 & 45.8$\pm$9.0 & 69.9$\pm$4.2 & \underline{52.2$\pm$3.0} & 79.6$\pm$10.4 & 64.0$\pm$3.2 & 65.3$\pm$2.6 & 64.6$\pm$3.1 \\
            T-TIME          & 69.3$\pm$3.7 & 57.8$\pm$1.8 & 76.7$\pm$6.9 & 45.4$\pm$8.6 & 70.8$\pm$3.3 & 51.3$\pm$3.0 & 79.4$\pm$10.8 & 64.9$\pm$1.4 & 65.3$\pm$1.8 & 64.5$\pm$2.3 \\
            EuHy            & 25.5$\pm$0.2 & 25.0$\pm$0.0 & 28.5$\pm$5.4 & 25.0$\pm$0.3 & 24.7$\pm$0.6 & 25.5$\pm$0.8 & 25.0$\pm$0.0 & 25.0$\pm$0.0 & 29.3$\pm$7.4 & 25.9$\pm$0.7 \\
            MEuHy           & 35.5$\pm$3.3 & 29.9$\pm$3.1 & 45.8$\pm$7.6 & 47.6$\pm$5.7 & 70.1$\pm$3.1 & 39.6$\pm$1.0 & 35.8$\pm$0.3 & 34.8$\pm$7.3 & 52.5$\pm$7.7 & 43.5$\pm$1.6 \\
            MRieHy          & \textbf{74.4$\pm$2.6} & \textbf{60.3$\pm$2.9} & \underline{83.0$\pm$2.8} & \textbf{57.6$\pm$3.5} & 71.4$\pm$2.0 & \textbf{54.6$\pm$4.5} & \textbf{85.8$\pm$2.7} & \textbf{77.3$\pm$2.4} & \textbf{76.6$\pm$3.8} & \textbf{71.2$\pm$0.9}\\
            \bottomrule
        \end{tabular}
        }
        \label{tab:bci}
    \end{subtable}
    \caption{Comparison between MRieHy and the baseline methods on three datasets}
    \label{tab:main}
\end{table*}

\subsection{Main Experimental Results}
Table \ref{tab:main} compares MRieHy against 10 baselines for online test-time adaptation on the ECoG128, Stieger2021, and BCI Competition IV 2a datasets.

On the ECoG128 dataset, MRieHy achieves the highest average accuracy of 64.1$\pm$0.9\% on test Days 6--8, surpassing the second-best method, BaseNet+RieMDM (62.7$\pm$0.8\%), by 1.4\% and outperforming other test-time adaptation methods by 1.9\%--6.1\%. EuHy (45.7$\pm$5.6\%) and MEuHy (56.5$\pm$2.3\%) perform drastically worse, highlighting the ineffectiveness of Euclidean cosine similarity and Euclidean alignment for SPD covariance matrices.

On the Stieger2021 dataset, MRieHy again achieves the highest average accuracy (54.1$\pm$0.4\%), though it leads CoTTA (53.2$\pm$0.5\%) by only 0.9\%. Its advantages become clear upon closer inspection: it ranks first on four of the seven subjects---more than any other method---while remaining competitive on the rest. This consistent, subject-level robustness demonstrates MRieHy's effectiveness as a test-time adaptation method.

On the BCI Competition IV 2a dataset, MRieHy continues to lead with an average accuracy of 71.2$\pm$0.9\% over 9 subjects, surpassing BaseNet (68.2$\pm$1.4\%), BaseNet+RieMDM (68.6$\pm$1.3\%), and RieMDM (58.6$\pm$0.2\%) by 2.6\%--12.6\%, and outperforming the other test-time adaptation methods by 6.6\%--11.1\%. EuHy (25.9$\pm$0.7\%) and MEuHy (43.5$\pm$1.6\%) again severely underperform.

MRieHy's consistent outperformance confirms its core advantages: Riemannian metric-based similarity and alignment better capture SPD manifold properties, and multi-feature fusion (deep + covariance) via hypergraphs leverages complementary information more effectively than the simple ensemble of BaseNet and RieMDM.

\subsection{Ablation Study}
We conducted an ablation study on all datasets to analyze the individual components of the MRieHy framework. Table \ref{tab:ablation} details the results for different alignment methods, feature combinations, and similarity strategies.

\textbf{Riemannian Similarity Outperforms Euclidean Methods.} Our results demonstrate a clear advantage of Riemannian geometry-based similarity measures over Euclidean methods. On all three datasets, Riemannian methods (TanCos, GauRie, RieDM, TanDM) consistently outperform Euclidean approaches (Cos, EuDM). While performance differences among Riemannian methods remain relatively small (within 3\%), suggesting that properly leveraging the SPD manifold geometry matters more than the specific Riemannian similarity implementation.

\textbf{Riemannian Alignment Better Preserves Covariance Geometry.} Comparing the 6th and 7th rows of Table \ref{tab:ablation} reveals the importance of Riemannian alignment. When replacing Riemannian alignment with Euclidean alignment, performance drops by 1.1\%--9.4\% on three datasets. This decrease confirms that Riemannian alignment better preserves the geometric structure of covariance matrices across different days, effectively mitigating distribution shifts in online test-time adaptation.

\textbf{Multi-feature Fusion Delivers Critical Gains.} Across all three datasets, the full MRieHy model consistently outperforms versions that rely on only one feature type. Replacing multi-feature fusion with covariance features alone or deep features alone causes accuracy drops ranging from 4.2\% to 15.8\%. This consistent finding demonstrates that both feature types contribute complementary information essential for robust decoding, and their fusion yields substantially better cross-day generalization than either feature alone.
\begin{table}
    \centering
    \small
    \begin{tabular}{l|c|c|c|c|c|c}
        \toprule
        Align    & Co         & Sim         & Deep       & ECoG128     & BCI Comp & Stieger2021\\
        \midrule
        Rie      & \checkmark & Cos         & \checkmark & 59.9$\pm$2.2 & 45.1$\pm$1.2 & 52.0$\pm$0.3\\
        Rie      & \checkmark & TanCos      & \checkmark & \underline{64.1$\pm$1.0} & \textbf{71.2$\pm$0.9} & \textbf{54.1$\pm$0.3}\\
        Rie      & \checkmark & GauRie      & \checkmark & 63.3$\pm$0.9 & 68.8$\pm$1.5 & 52.5$\pm$0.4\\
        Rie      & \checkmark & EuDM        & \checkmark & 59.6$\pm$2.1 & 44.8$\pm$1.1 & 52.0$\pm$0.2\\
        Rie      & \checkmark & RieDM       & \checkmark & 63.3$\pm$0.9 & 68.8$\pm$1.5 & 52.5$\pm$0.3\\
        Rie      & \checkmark & TanDM       & \checkmark & \textbf{64.1$\pm$0.9} & \textbf{71.2$\pm$0.9} & \underline{54.1$\pm$0.4}\\
        Eu       & \checkmark & TanDM       & \checkmark & 63.0$\pm$0.4 & 70.1$\pm$1.2 & 44.7$\pm$0.1\\
        Rie      & \checkmark & TanDM       & $\times$   & 59.9$\pm$0.1 & 55.8$\pm$1.0 & 46.6$\pm$0.1\\
        Rie      & $\times$   & -           & \checkmark & 59.8$\pm$0.9 & 66.3$\pm$1.0 & 51.1$\pm$0.1\\
        \bottomrule
    \end{tabular}
    \caption{Ablation study of different alignment methods, features (Co\&Deep), and similarity computation (Sim) for covariance matrix}
    \label{tab:ablation}
\end{table}
\subsection{Number of Days Included in a Hyperedge}
\begin{figure}[htb]
    \centering
    % �� 核心修复：一次性全局设置所有子图标题左对齐（不破坏任何间距）
    \captionsetup[subfigure]{justification=raggedright,singlelinecheck=false}
    
    % 统一宽度 + 标准等间距排版
    \begin{subfigure}{0.24\textwidth}
        \centering
        \includegraphics[width=\linewidth]{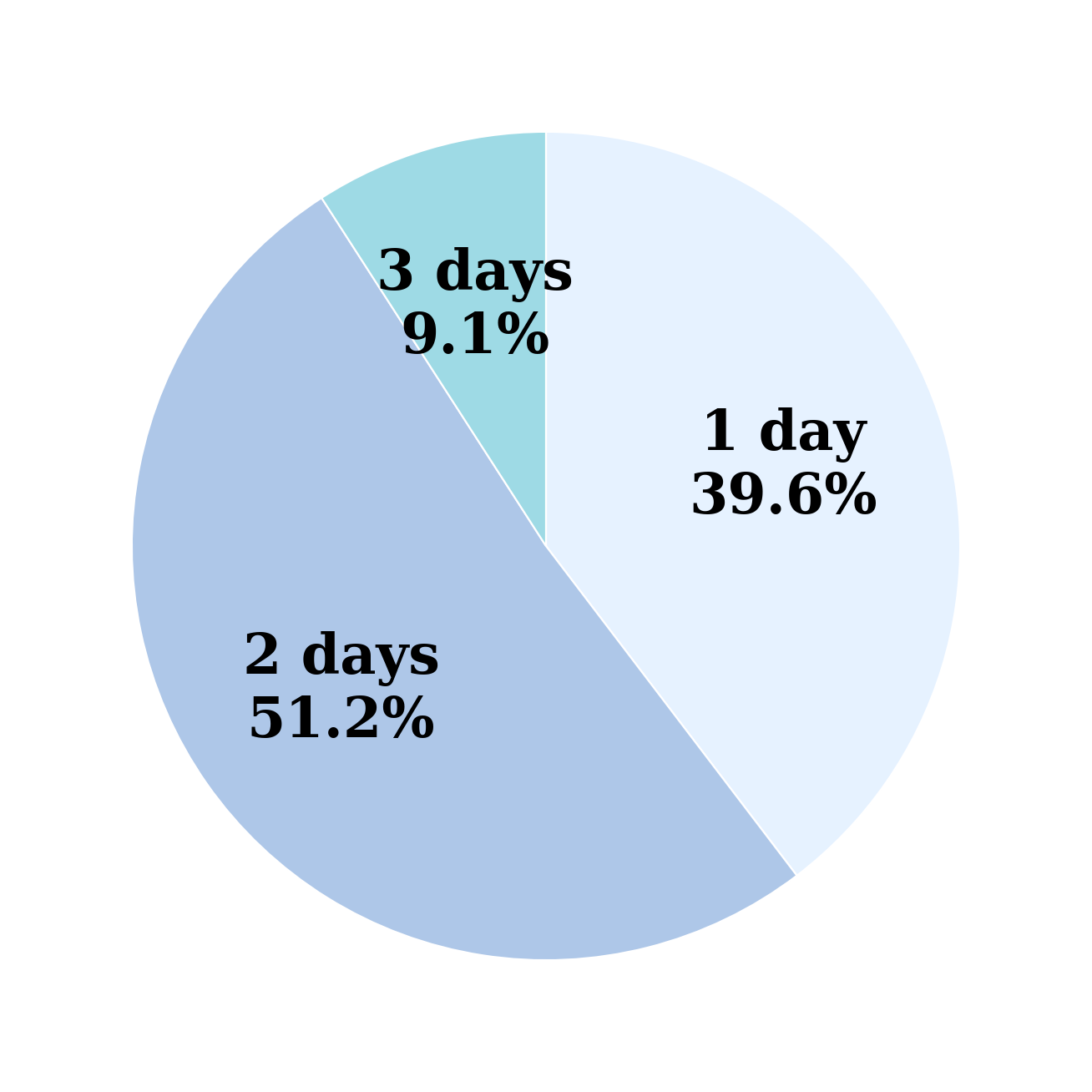}
        \caption{Dataset: ECoG128  Sim: Cos}
    \end{subfigure}
    \hfill  % 标准均匀间距
    \begin{subfigure}{0.24\textwidth}
        \centering
        \includegraphics[width=\linewidth]{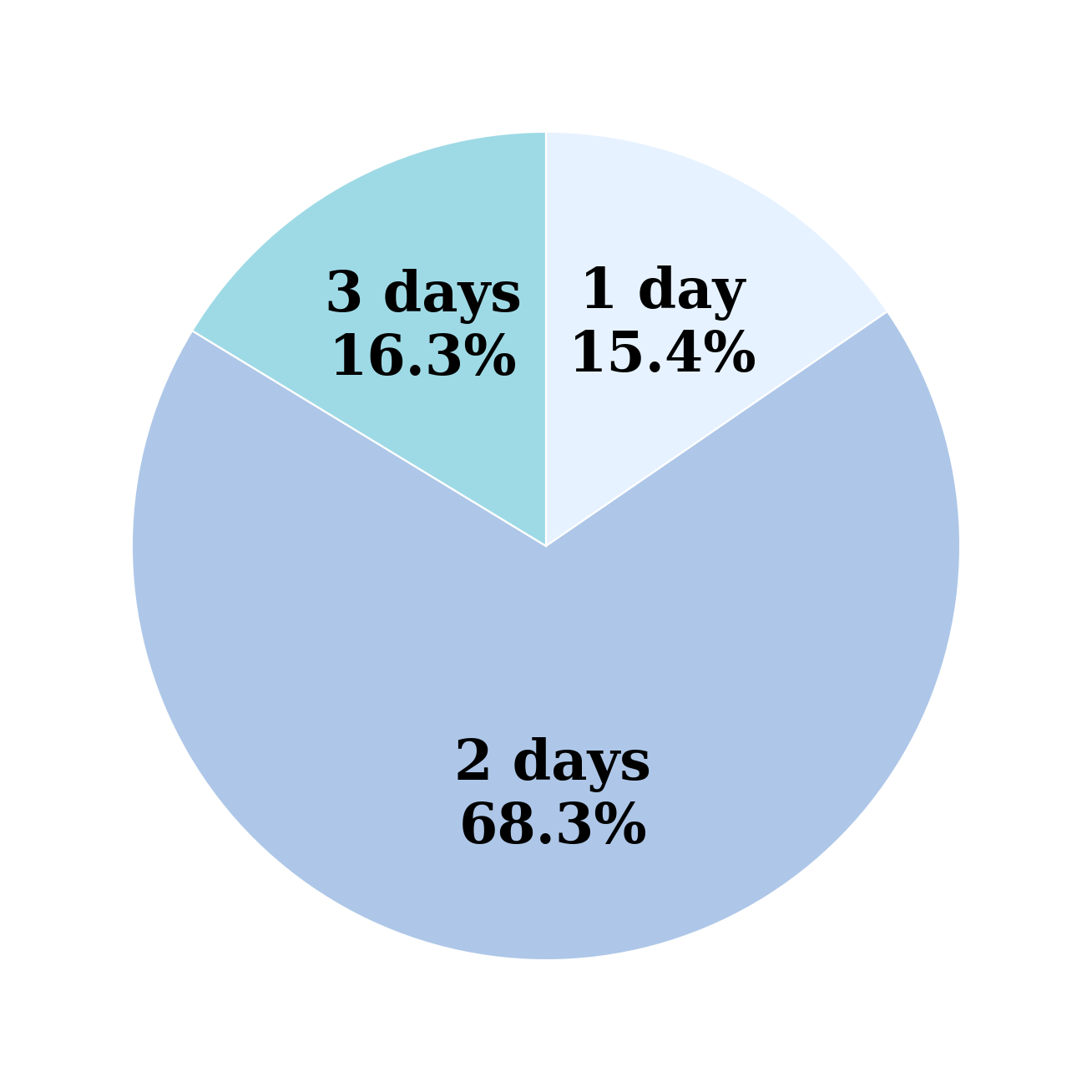}
        \caption{Dataset: ECoG128 Sim: TanDM}
    \end{subfigure}
    \hfill  % 标准均匀间距
    \begin{subfigure}{0.24\textwidth}
        \centering
        \includegraphics[width=\linewidth]{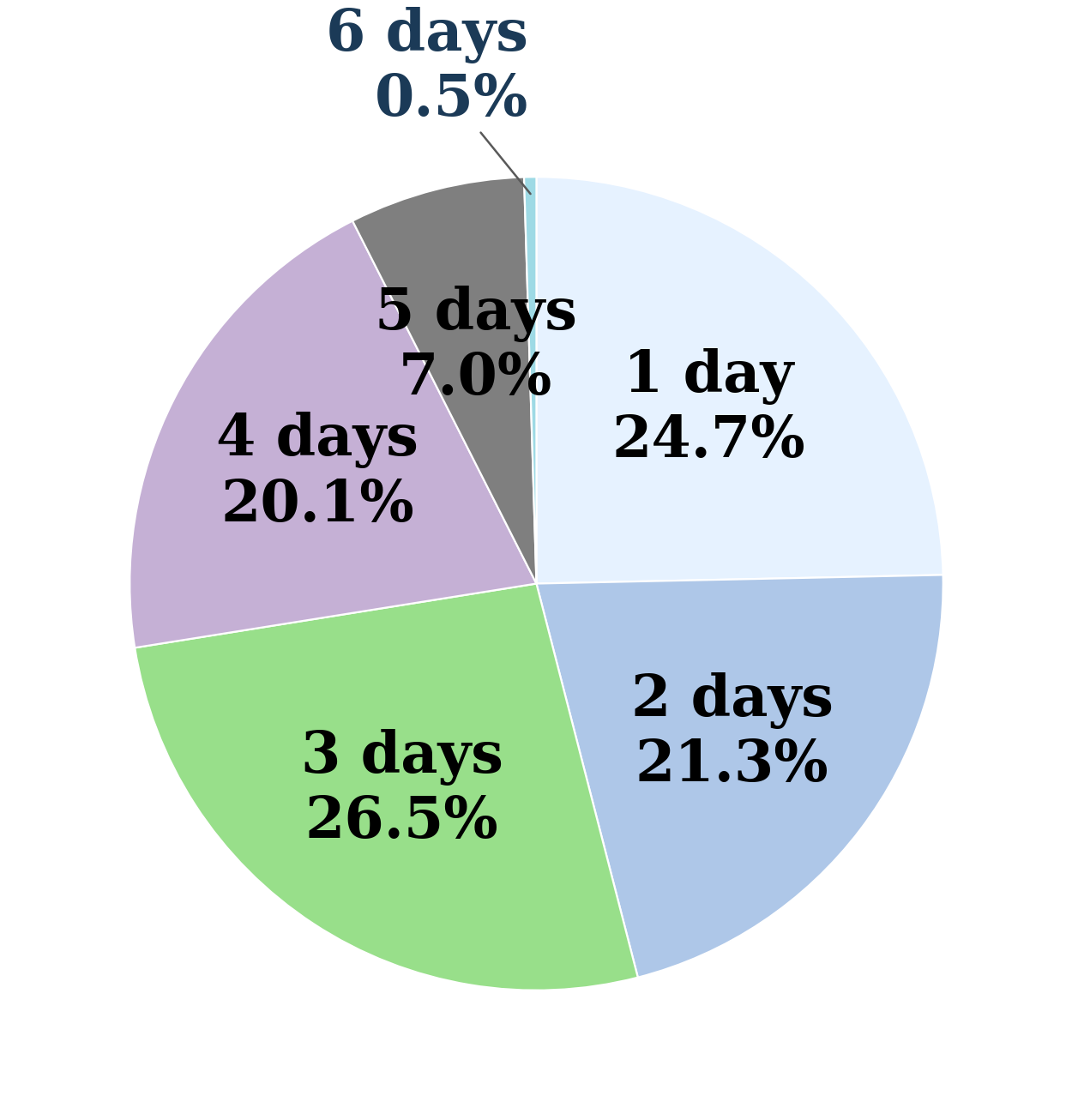}
        \caption{Dataset: Stieger2021 Subj:1 Sim: Cos}
    \end{subfigure}
    \hfill  % 标准均匀间距
    \begin{subfigure}{0.24\textwidth}
        \centering
        \includegraphics[width=\linewidth]{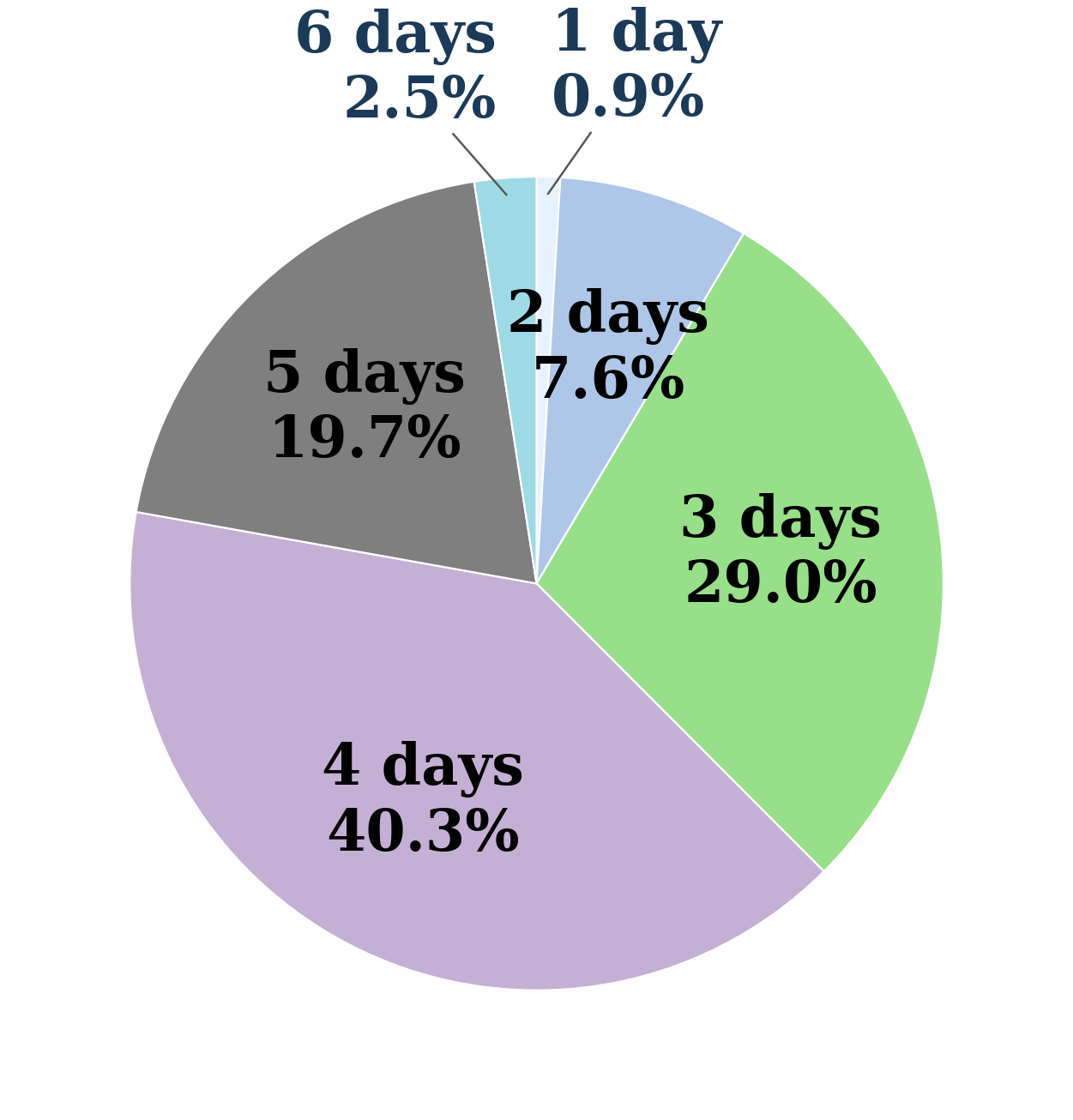}
        \caption{Dataset: Stieger2021 Subj:1 Sim: TanDM}
    \end{subfigure}
    
    \caption{Percentage of distinct days per hyperedge in the covariance‑based hypergraph component}
    \label{fig:pie_chart}
\end{figure}

We count the distinct days per hyperedge in the covariance‑based hypergraph component. On both ECoG128 and Stieger2021, hypergraphs built with TanDM contain more distinct days per hyperedge than those using cosine similarity, while the number of samples per hyperedge stays constant (Figure~\ref{fig:pie_chart}, Figure~\ref{fig:pie_chart_s}). This suggests that Riemannian distance‑based similarity better aligns samples across days, likely explaining MRieHy’s superior performance.

\begin{figure}[htb]
    \centering
    \begin{subfigure}{0.4\textwidth}
        \centering
        \includegraphics[width=\linewidth]{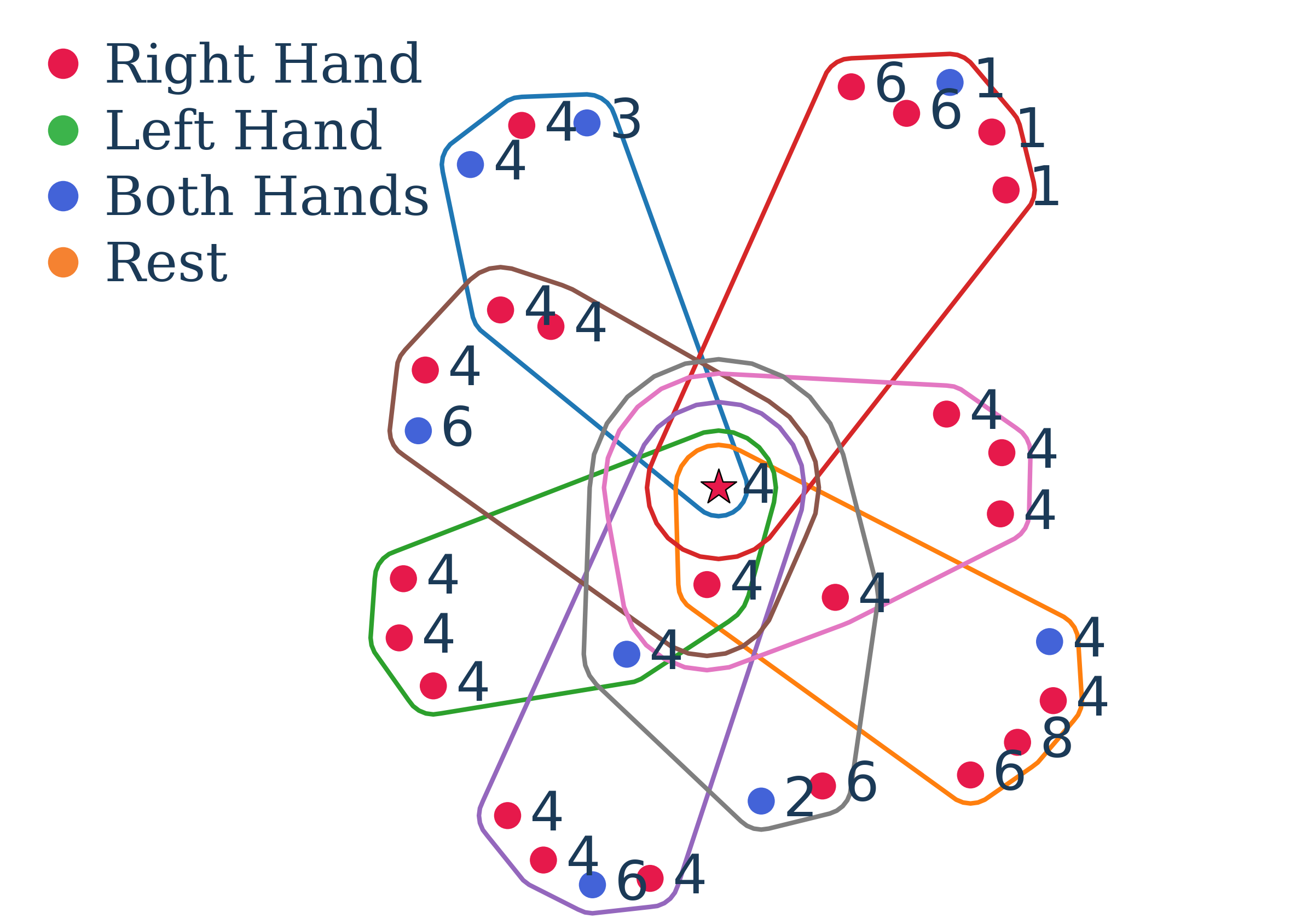}
        %\caption{Hypergraph built with cosine similarity}
        \label{fig:ego_vis1}
    \end{subfigure}
    \hfill
    \begin{subfigure}{0.4\textwidth}
        \centering
        \includegraphics[width=\linewidth]{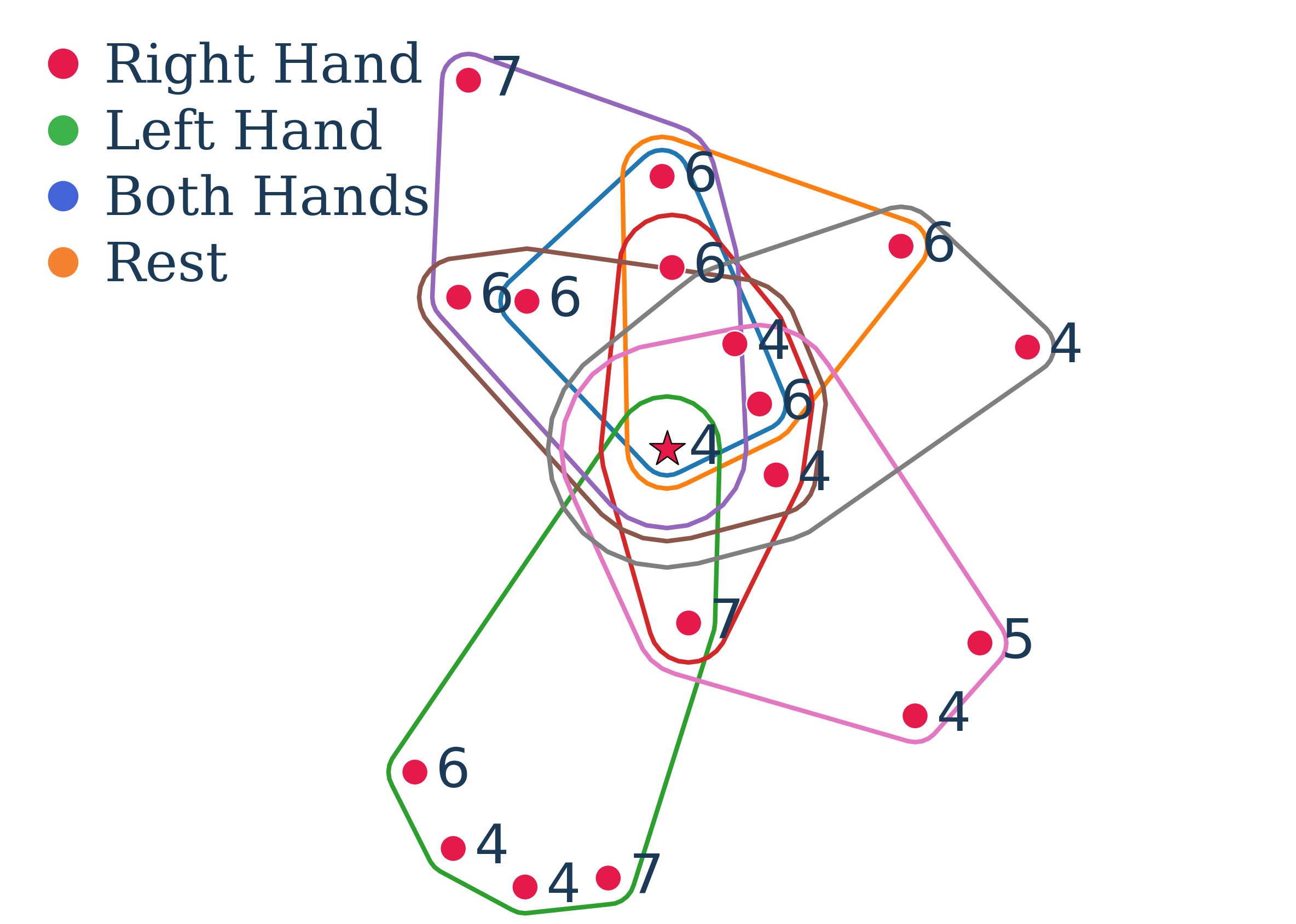}
        %\caption{Riemannian distance-based similarity}
        \label{fig:ego_vis2}
    \end{subfigure}
    \caption{Neighbor vertices and incident hyperedges for a sample from Stieger2021 dataset Subject 1 in the covariance-based hypergraph component built with cosine similarity (left) and with TanDM (right). Numbers indicate the day each vertex is from.}
    \label{fig:ego_vis}
\end{figure}

Figure \ref{fig:ego_vis} shows the neighbor vertices and incident hyperedges for the same sample, taken from Subject 1 of the Stieger2021 dataset, in the covariance‑based hypergraph component built with cosine similarity and with TanDM. In the cosine‑similarity case 19 of 31 neighbor vertices belong to Day 4---the same day as the target vertex. The TanDM case contains only 6 of 17 neighbor vertices from Day 4; the remaining vertices are same‑class (Right Hand) samples from various days. This intuitively supports that Riemannian distance‑based similarity helps hyperedges aggregate same‑class vertices across different days.

\subsection{Impact of the Number of Training Days}\label{subsec:n_train_d}

\begin{figure}[h]
    \centering
    % 子图容器加宽，给标题足够空间
    \begin{subfigure}{0.48\linewidth}
        \centering
        % �� 关键：固定高度，保持比例，仅宽度自适应增加
        \includegraphics[height=3.2cm, keepaspectratio]{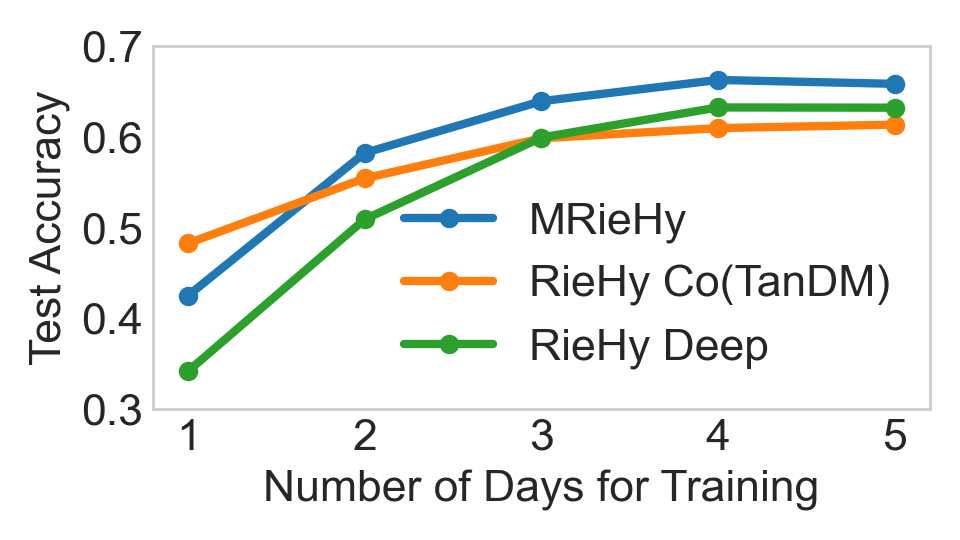}
        \caption{ECoG128 dataset}
    \end{subfigure}
    \hfill
    \begin{subfigure}{0.48\linewidth}
        \centering
        % 和左图高度完全一致，宽度自动增加
        \includegraphics[height=3.2cm, keepaspectratio]{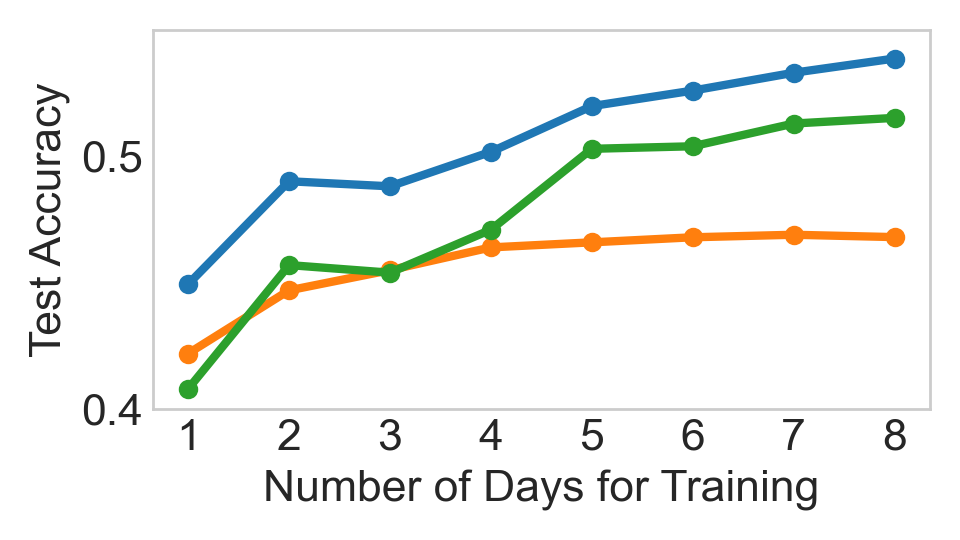}
        \caption{Average of 7 subjects from Stieger2021 dataset} % 一行显示！
    \end{subfigure}
    \caption{Impact of training day count on MRieHy and component hypergraph performance. }
    \label{fig:accuracy_by_day}
\end{figure}

We study how training days affect test accuracy for MRieHy and its two components, as shown in Figure \ref{fig:accuracy_by_day}. On ECoG128, MRieHy’s accuracy rises markedly from one to three training days and then plateaus from four to five days; on Stieger2021, it improves steadily as more training days are added.

The two feature components exhibit distinct learning patterns. The deep feature‑based component performs poorly with limited data but improves considerably when more data become available, whereas the covariance‑based component, which relies on Riemannian tangent space distance to mean similarity, remains relatively stable across training days. This aligns with known characteristics of BaseNet’s data dependence and the stability of Riemannian MDM \cite{Wimpff2025TailoringDL}. These results confirm that our tangent space similarity inherits Riemannian geometry's stability, and that MRieHy's multi-feature architecture effectively combines both feature types' complementary strengths: covariance representations are especially valuable with limited training data, while deep features become advantageous with larger datasets.

\subsection{Parameter Sensitivity}

\begin{figure}[h]
    \centering
    \begin{subfigure}{0.32\textwidth}
        \centering
        \includegraphics[width=\linewidth]{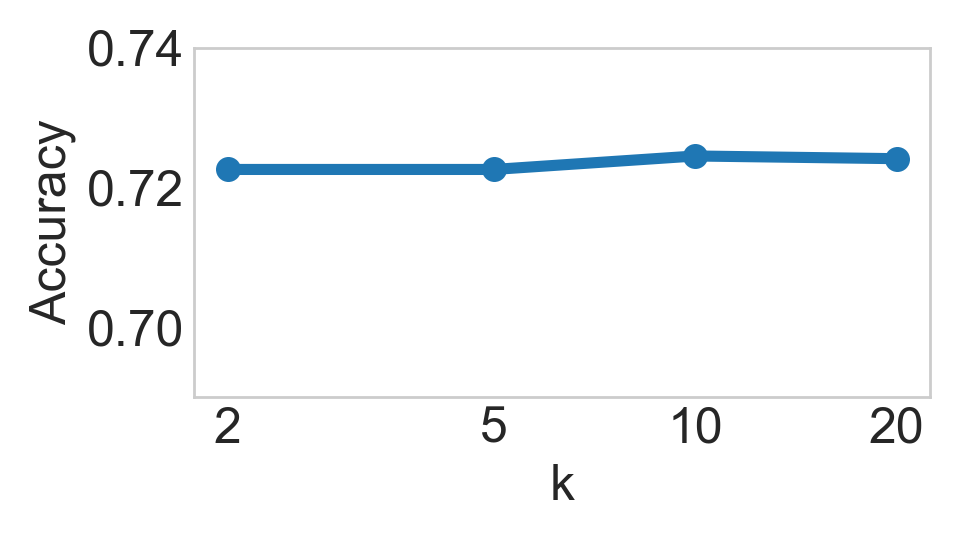}
        % \caption{Neighbor number $k$}
        \label{fig:sub1}
    \end{subfigure}
    \hfill
    \begin{subfigure}{0.32\textwidth}
        \centering
        \includegraphics[width=\linewidth]{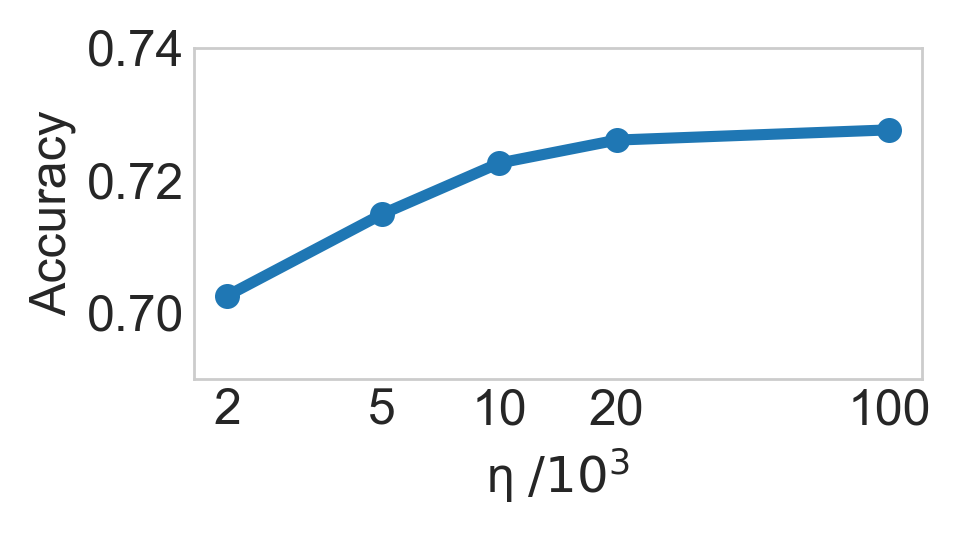}
        % \caption{Regularization parameter $\eta$}
        \label{fig:sub2}
    \end{subfigure}
    \hfill
    \begin{subfigure}{0.32\textwidth}
        \centering
        \includegraphics[width=\linewidth]{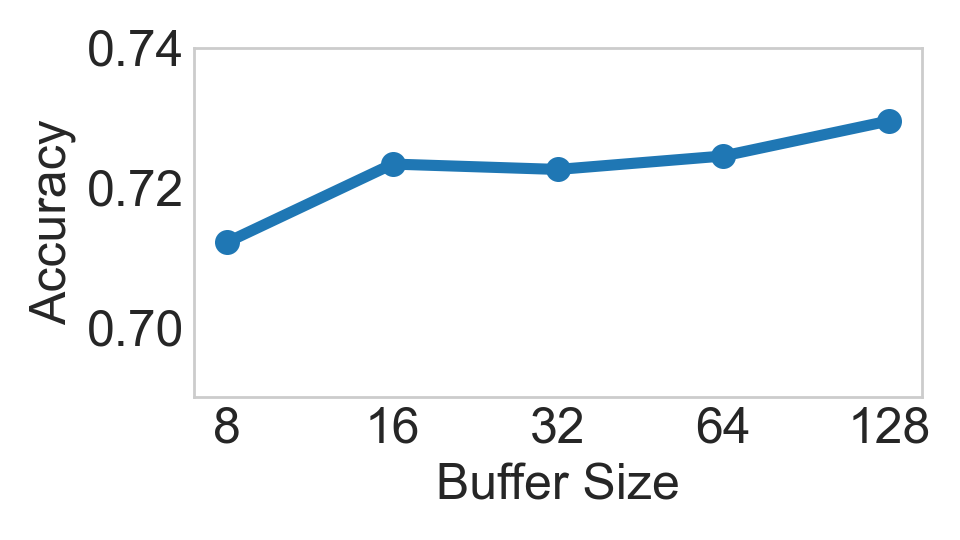}
        % \caption{Buffer size}
        \label{fig:sub3}
    \end{subfigure}
    \hfill
    \begin{subfigure}{0.32\textwidth}
        \centering
        \includegraphics[width=\linewidth]{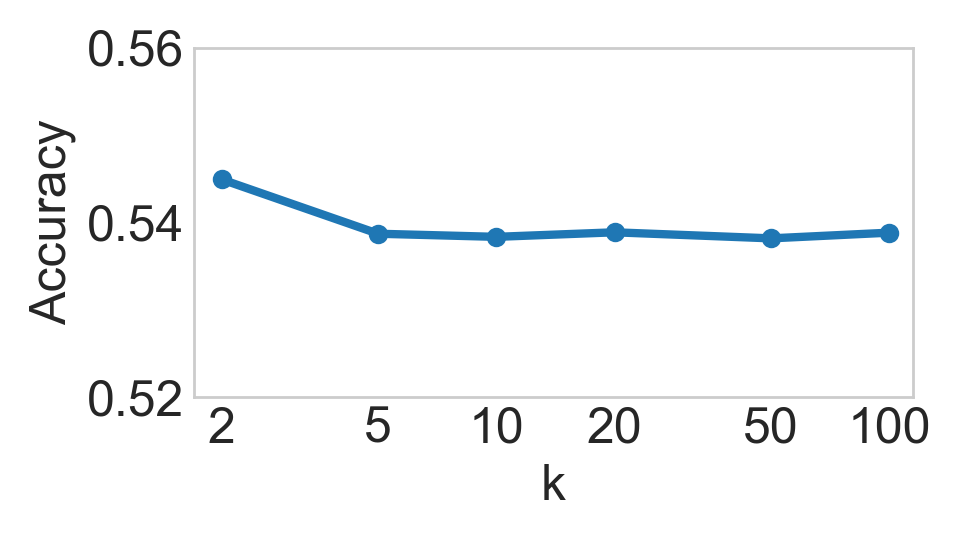}
        % \caption{Buffer size}
        \label{fig:sub3}
    \end{subfigure}     
    \hfill
    \begin{subfigure}{0.32\textwidth}
        \centering
        \includegraphics[width=\linewidth]{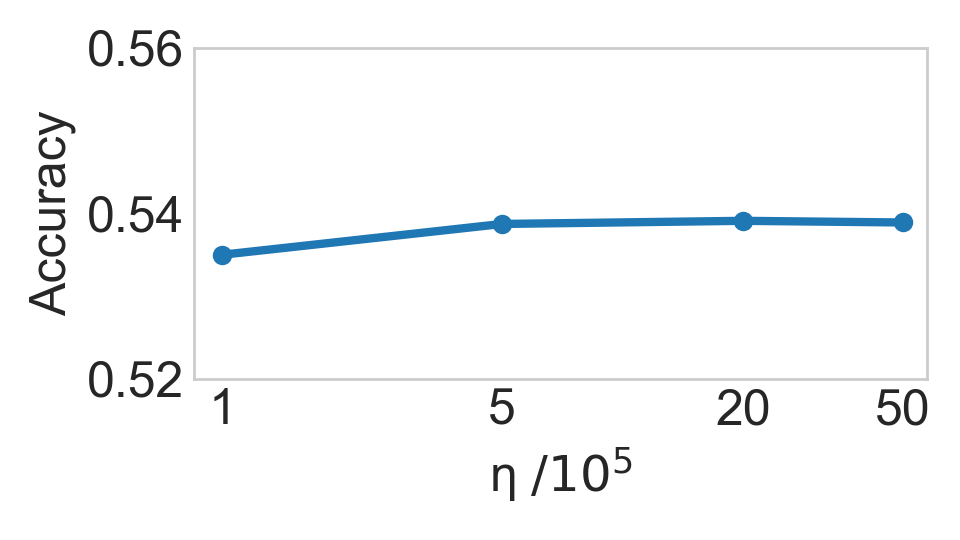}
        % \caption{Buffer size}
        \label{fig:sub3}
    \end{subfigure}     
    \hfill
    \begin{subfigure}{0.32\textwidth}
        \centering
        \includegraphics[width=\linewidth]{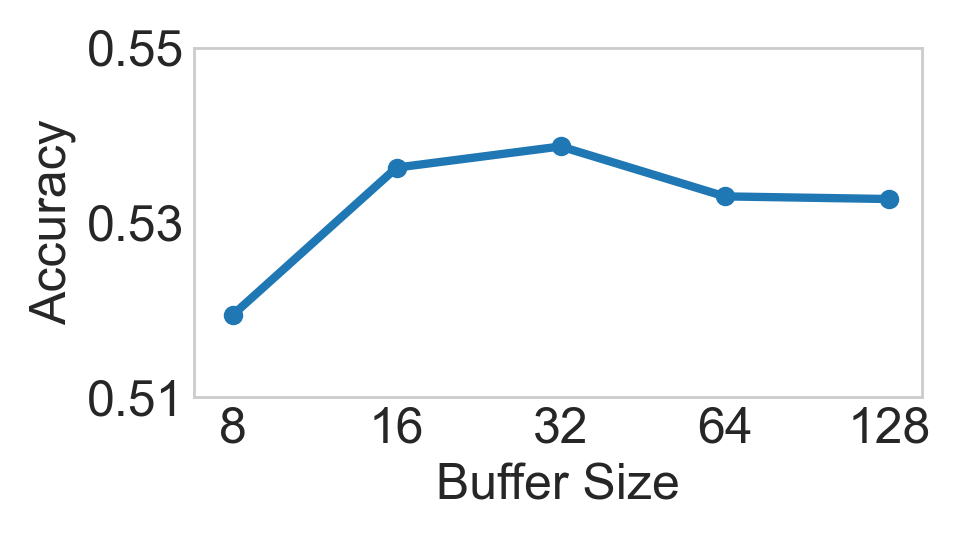}
        % \caption{Buffer size}
        \label{fig:sub3}
    \end{subfigure}    
    \caption{Effect of $k$, $\eta$, and buffer size on MRieHy’s average test accuracy across 9 subjects on BCI Competition IV 2a (upper row) and 7 subjects on Stieger2021 (bottom row).}
    \label{fig:param_sensitivity}
\end{figure}

We further evaluate MRieHy’s sensitivity to three key hyperparameters: the number of neighbors $k$ for hyperedge generation, the regularization weight $\eta$ for multi‑hypergraph fusion, and the online buffer size. Experiments were conducted on the BCI Competition IV 2a dataset (Figure \ref{fig:param_sensitivity} upper row) with default values $k=2$, $\eta=10000$, buffer size $=32$, and on the Stieger2021 dataset (Figure \ref{fig:param_sensitivity} bottom row)with defaults $k=5$, $\eta=500{,}000$, buffer size $=32$. In each analysis, one hyperparameter is varied while the others remain fixed.

As shown in Figure \ref{fig:param_sensitivity} left column, MRieHy’s accuracy stays relatively stable across different values of $k$ on both datasets. For $\eta$, accuracy increases slightly as $\eta$ grows on the BCI Competition dataset (Figure \ref{fig:param_sensitivity} upper middle) but remains stable on Stieger2021 (Figure \ref{fig:param_sensitivity} bottom middle). Finally, Figure \ref{fig:param_sensitivity} right column indicates that larger buffer sizes yield a small accuracy gain on the BCI Competition dataset, whereas on Stieger2021 accuracy increases marginally and then plateaus as the buffer grows. These results suggest that optimal hyperparameter combinations vary across datasets, yet overall performance remains robust to changes in these hyperparameters.

\section{Conclusion}\label{sec:conclusion}

In conclusion, this paper presented the Multi-feature Riemannian Hypergraph (MRieHy), a novel framework for online test-time adaptation of MI-BCIs. By combining a Riemannian-distance-based hypergraph over covariance matrices with a cosine-similarity-based hypergraph over deep features, MRieHy effectively mitigates cross-day performance degradation without labeled test-day data. A limitation lies in the buffer sampling, which lacks stability analysis and outlier rejection mechanisms. Future work will explore more robust buffer management with anomaly detection for long‑term plug‑and‑play BCI use.

% \begin{ack}
% We acknowledge the support from funding agencies and research institutions involved in this work.
% \end{ack}

\small

\bibliography{MRieHy}
\bibliographystyle{plain}

\appendix
\counterwithin{figure}{section}
\counterwithin{table}{section}

\section{Multi-feature Hypergraph Learning Details}
Let $\mathbf{Z}=[\mathbf{f}_1,\mathbf{f}_2,\cdots,\mathbf{f}_n]$ be the concatenation of all sample features, where $n$ is the total sample number for training. $\mathbf{Z}\in\mathbb{R}^{C\cdot C\times n}$ when using the covariance matrix as feature, and $\mathbf{Z}\in\mathbb{R}^{d\times n}$ when using the deep feature. Let $\mathbf{y}_i=\text{onehot}(y_i)\in\mathbb{R}^N$, and let $\mathbf{Y}=[\mathbf{y}_1, \mathbf{y}_2,\cdots,\mathbf{y}_n]^\top\in\mathbb{R}^{n\times N}$ denote the label matrix for all training samples, where $y_i$ is the label of sample $i$. Following \cite{Zhang2018InductiveML}, the goal of hypergraph learning is to learn a regularized matrix $\mathbf{M}$ to project the feature matrix $\mathbf{Z}$ to the label space to discriminate different categories.

To learn the projection matrix $\mathbf{M}$ for a single hypergraph, a cost function $\mathbf{\Psi}$ is introduced, aggregating the hypergraph Laplacian regularizer $\mathbf{\Omega}(\mathbf{M})$, the empirical loss $\mathcal{R}_{emp}(M)$, and a regularization term on $\mathbf{M}$ given by $\mathbf{\Phi}(\mathbf{M})$:
\begin{equation}
    \mathbf{\Psi}=\{\mathbf{\Omega}(\mathbf{M})+\lambda\mathcal{R}_{emp}(M)+\mu\mathbf{\Phi}(\mathbf{M})\}
\label{eq:cost_func}
\end{equation}

The hypergraph Laplacian regularizer for $\mathbf{M}$ is under the assumption that vertices with strong connections should share similar labels. The hypergraph Laplacian regularizer is written as equation \eqref{eq:hypergraph_laplacian}, and it takes a quadratic form in $\mathbf{M}$.
\begin{equation}
\Omega(\mathbf{M}) = \frac{1}{2} \sum_{k=1}^{N} \sum_{e \in \mathcal{E}} \sum_{u,v \in \mathcal{V}} \frac{\mathbf{W}(e) \, \mathbf{H}(u, e) \, \mathbf{H}(v, e)}{\delta(e)} \vartheta = \operatorname{tr}\left( \mathbf{M}^{\mathrm{T}} \mathbf{Z} \Delta \mathbf{Z}^{\mathrm{T}} \mathbf{M} \right)
\label{eq:hypergraph_laplacian}
\end{equation}
where $\vartheta = \left( \frac{(\mathbf{Z}^{\mathrm{T}} \mathbf{M})(u,k)}{\sqrt{d(u)}} - \frac{(\mathbf{Z}^{\mathrm{T}} \mathbf{M})(v,k)}{\sqrt{d(v)}} \right)^2$, and $\Delta = \mathbf{I}-(\mathbf{D}^v)^{-\frac{1}{2}} \mathbf{H} \mathbf{W} (\mathbf{D}^e)^{-1} \mathbf{H}^\top (\mathbf{D}^v)^{-\frac{1}{2}}$.

The empirical loss associated with $\mathbf{M}$ is expressed as
\begin{equation}
\mathcal{R}_{emp}(\mathbf{M}) = ||\mathbf{Z}^{\mathrm{T}} \mathbf{M} - \mathbf{Y}||^2
\end{equation}

To mitigate overfitting in $\mathbf{M}$ while encouraging row-wise sparsity that highlights informative features, the $\ell_{2,1}$ norm regularizer is employed, defined by
\begin{equation}
\Phi(\mathbf{M}) = ||\mathbf{M}||_{2,1}
\end{equation}

Putting these pieces together, the learning objective over the hypergraph is formulated as
\begin{equation}
\arg\min_{\mathbf{M}}\left\{ \operatorname{tr}\left( \mathbf{M}^\top \mathbf{Z} \Delta \mathbf{Z}^\top \mathbf{M} \right) + \lambda || \mathbf{Z}^{\mathrm{T}} \mathbf{M} - \mathbf{Y} ||^2 + \mu || \mathbf{M} ||_{2,1} \right\}
\label{eq:learning_objective}
\end{equation}

Equation \eqref{eq:learning_objective} can be solved by an iteration process. For the detailed derivation and solution steps, we refer the reader to \cite{Zhang2018InductiveML}.

In the context of multi-feature hypergraphs, the cost function $\overline{\Psi}$ for learning the projection matrices $\mathbf{M}_h$ from all hypergraphs comprises two components: the aggregated learning costs across individual hypergraphs and a regularization term applied to the combination weights $\boldsymbol{\omega}$:
\begin{equation}
\overline{\Psi} = \sum_{h=1}^{m} \omega_h \left\{ \Omega(\mathbf{M}_h) + \lambda \mathcal{R}_{emp}(\mathbf{M}_h) + \mu \Phi(\mathbf{M}_h) \right\} + \eta \Gamma(\boldsymbol{\omega})
\end{equation}

where $m$ denotes the number of hypergraphs (equaling 2 in this research), and $h$ indexes the specific hypergraph. In this formulation, $\Gamma(\boldsymbol{\omega})$ is defined as the $\ell_2$ norm of the modality weight vector:
\begin{equation}
\Gamma(\boldsymbol{\omega}) = \|\boldsymbol{\omega}\|^2
\end{equation}

This regularization term is designed to determine optimal modality-specific combination weights. The associated learning objective for the multi-hypergraph framework is formulated as:
\begin{equation}
\begin{aligned}
\arg\min_{\mathbf{M}_h, \boldsymbol{\omega} \geq 0} \left\{ \sum_{h=1}^{m} \omega_h \left\{ \Omega(\mathbf{M}_h) + \lambda \mathcal{R}_{emp}(\mathbf{M}_h) + \mu \Phi(\mathbf{M}_h) \right\} + \eta \Gamma(\boldsymbol{\omega}) \right\}, \text{s.t.} \sum_{h=1}^{m} \omega_h = 1
\end{aligned}
\label{eq:multi_hypergraph_objective}
\end{equation}

Equation \eqref{eq:multi_hypergraph_objective} decomposes into $m + 1$ separable subproblems, each corresponding to a distinct projection matrix $\mathbf{M}_h$ and the weight vector $\boldsymbol{\omega}$. Consequently, the optimization proceeds by first solving for each $\mathbf{M}_h$ independently with method similar to that for equation \eqref{eq:cost_func}, followed by optimizing $\boldsymbol{\omega}$ to integrate the modalities.

With learned $\mathbf{M}_h$, the solution of the combination weight $\omega_h$ is
\begin{equation}
\omega_h = \frac{1}{m} + \frac{\sum_{h=1}^{m} \Upsilon_h}{2m\eta} - \frac{\Upsilon_h}{2\eta}
\end{equation}

where $\Upsilon_h=\Omega(\mathbf{M}_h)+\lambda\mathcal{R}_{emp}(\mathbf{M}_h)+\mu\Phi(\mathbf{M}_h)$. For detailed derivation, we also refer readers to \cite{Zhang2018InductiveML}.

\section{Computational Cost}\label{sec:cost}
The experiments are conducted on NVIDIA GeForce RTX 3090 Ti GPU with 12th Gen Intel(R) Core(TM) i9-12900K CPU. The typical off-line training time of Multi-feature Riemannian Hypergraph is about 63ms per sample, and the on-line inference time is about 14ms per sample.

\section{Ethics Statement}\label{sec:ethics}
Experiments that contribute to this work were approved by IRB. All subjects consent to participate. All electrode locations are exclusively dictated by clinical considerations.

\section{Percentage of Distinct Days per Hyperedge for More Subjects}

\begin{figure}[h]
    \centering
    % �� 核心修复：一次性全局设置所有子图标题左对齐（不破坏任何间距）
    \captionsetup[subfigure]{justification=raggedright,singlelinecheck=false}
    
    % 统一宽度 + 标准等间距排版
    \begin{subfigure}{0.24\textwidth}
        \centering
        \includegraphics[width=\linewidth]{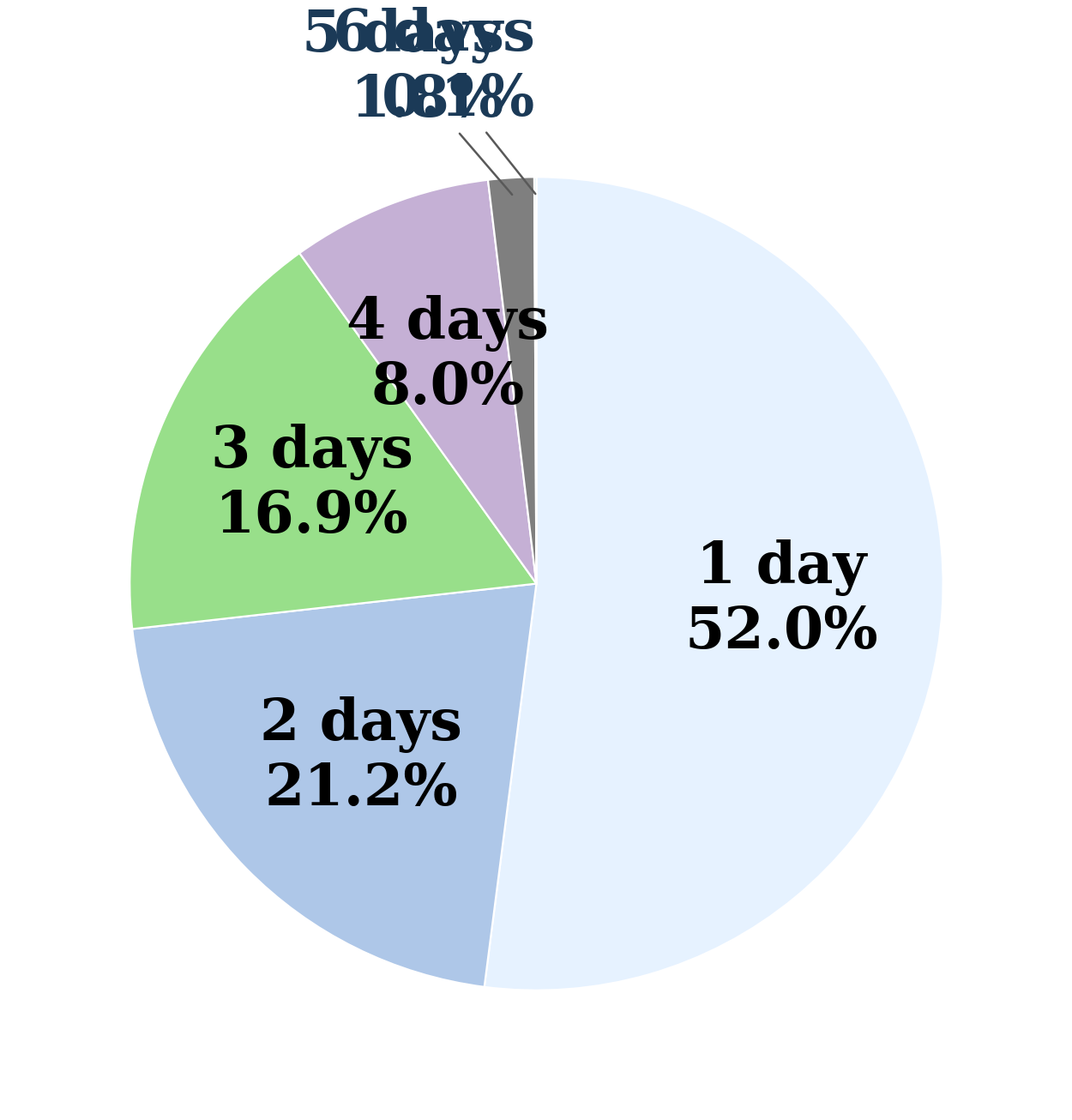}
        \caption{Dataset: Stieger2021 Subj:12 Sim: Cos}
    \end{subfigure}
    \hfill  % 标准均匀间距
    \begin{subfigure}{0.24\textwidth}
        \centering
        \includegraphics[width=\linewidth]{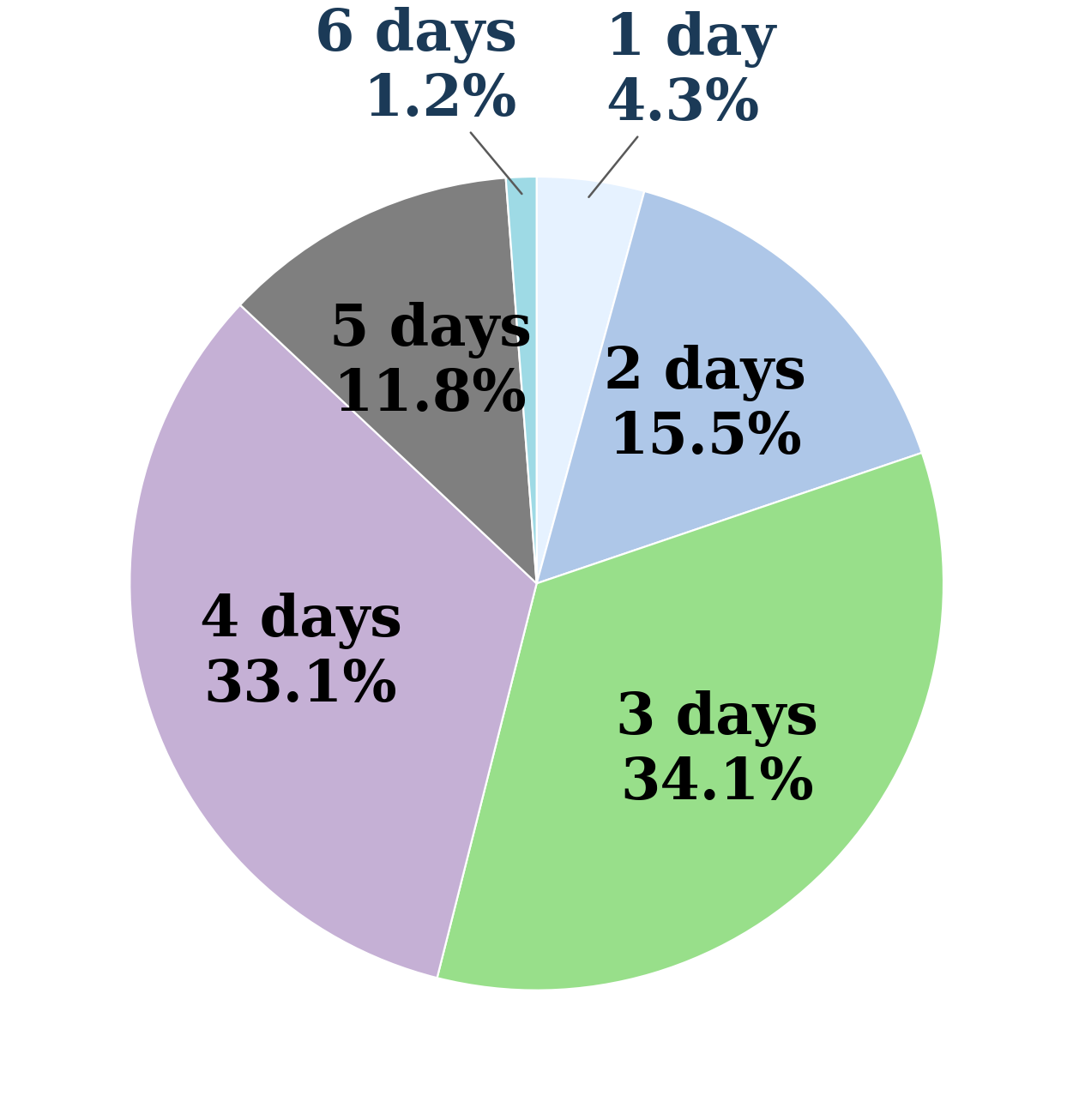}
        \caption{Dataset: Stieger2021 Subj:12 Sim: TanDM}
    \end{subfigure}
    \hfill  % 标准均匀间距
    \begin{subfigure}{0.24\textwidth}
        \centering
        \includegraphics[width=\linewidth]{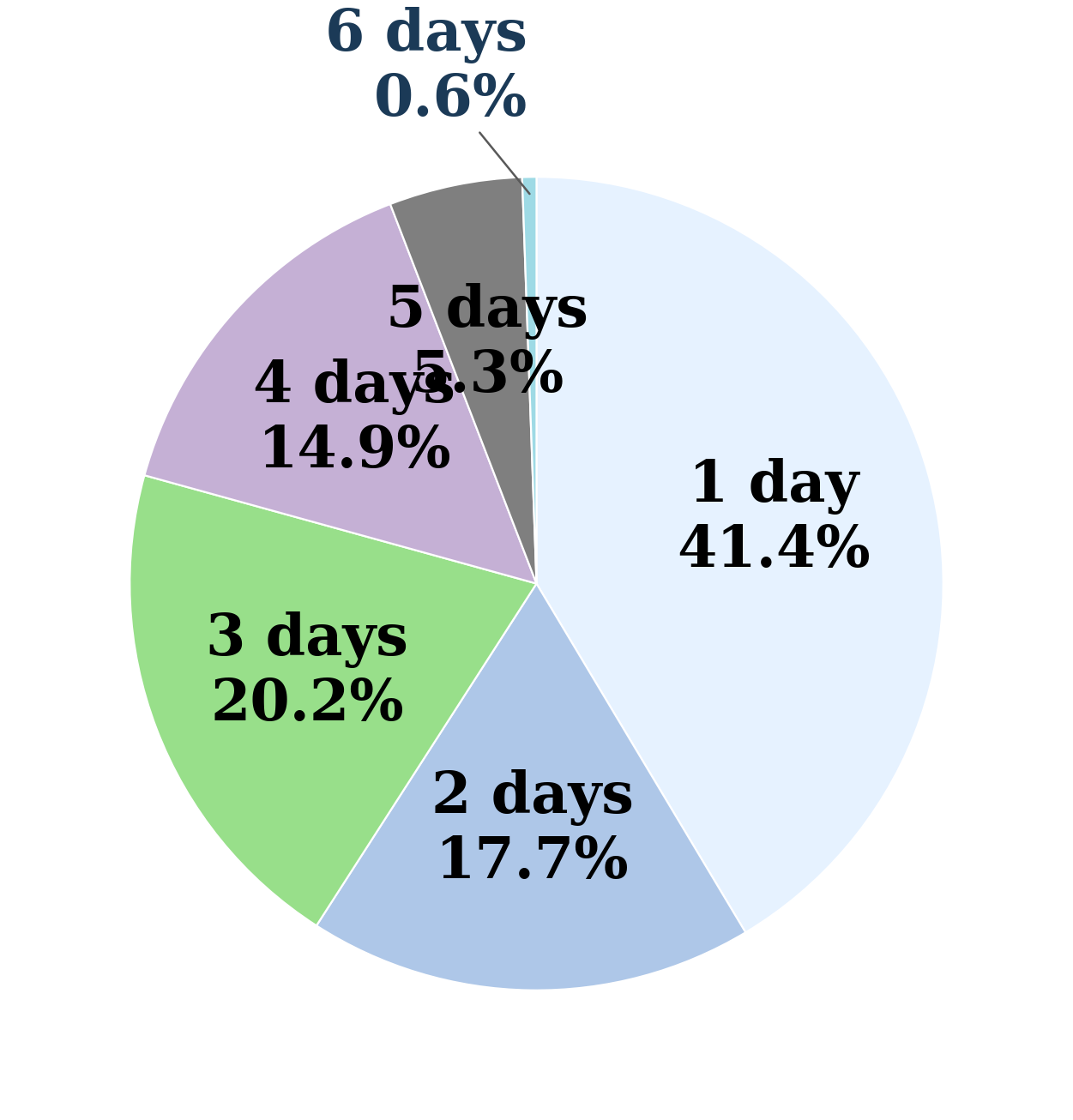}
        \caption{Dataset: Stieger2021 Subj:26 Sim: Cos}
    \end{subfigure}
    \hfill  % 标准均匀间距
    \begin{subfigure}{0.24\textwidth}
        \centering
        \includegraphics[width=\linewidth]{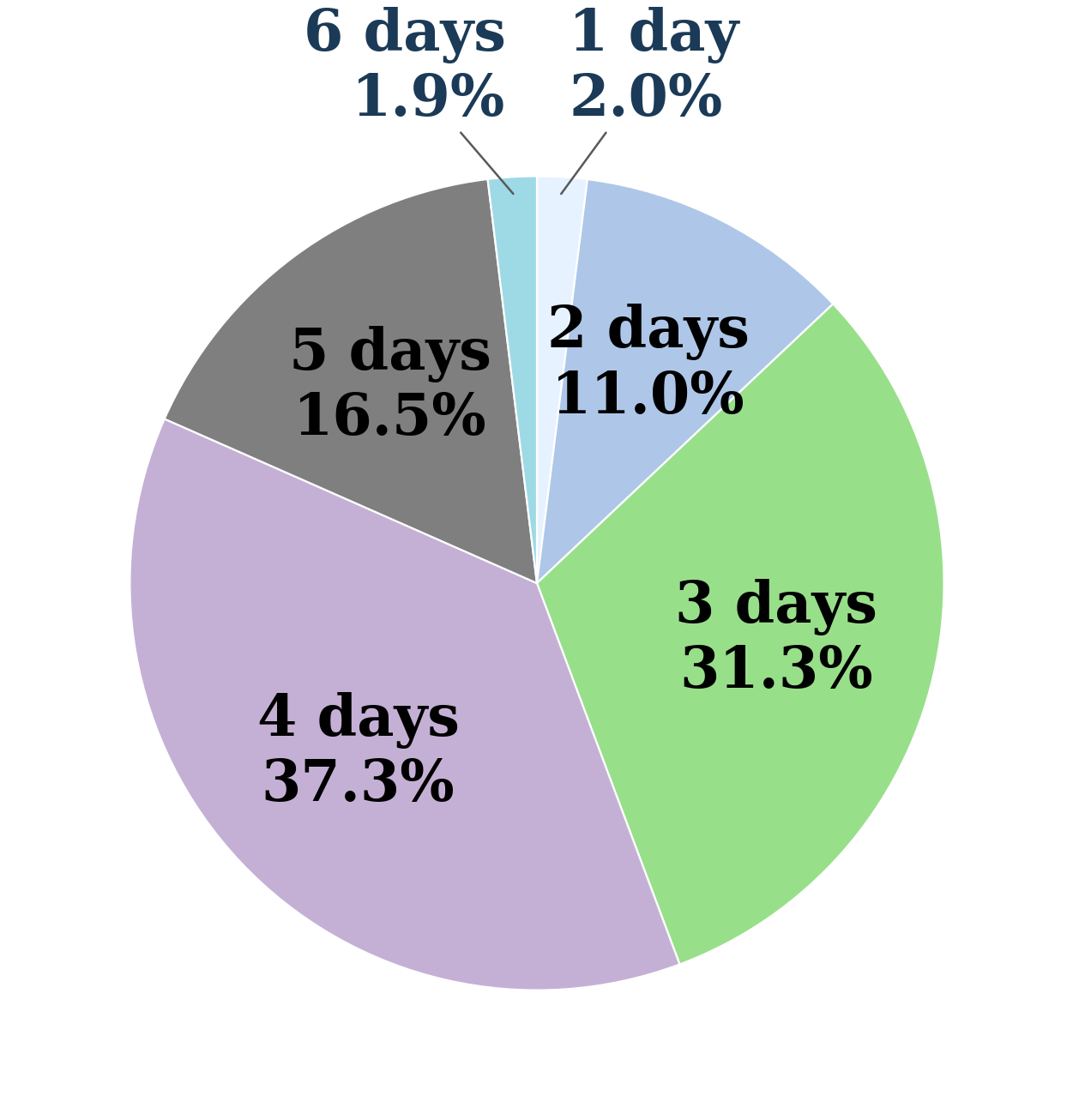}
        \caption{Dataset: Stieger2021 Subj:26 Sim: TanDM}
    \end{subfigure}
    \begin{subfigure}{0.24\textwidth}
        \centering
        \includegraphics[width=\linewidth]{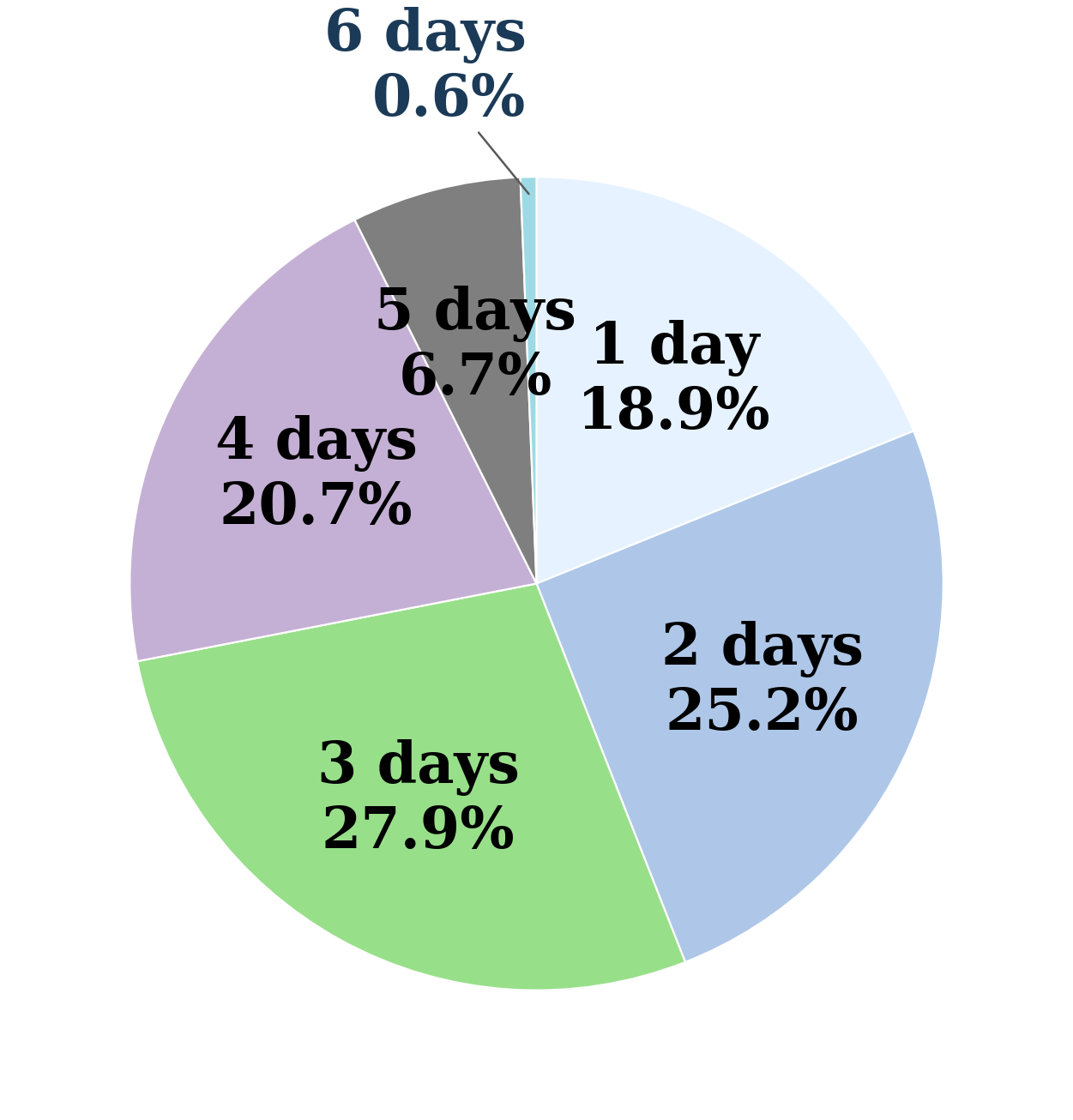}
        \caption{Dataset: Stieger2021 Subj:28 Sim: Cos}
    \end{subfigure}
    \begin{subfigure}{0.24\textwidth}
        \centering
        \includegraphics[width=\linewidth]{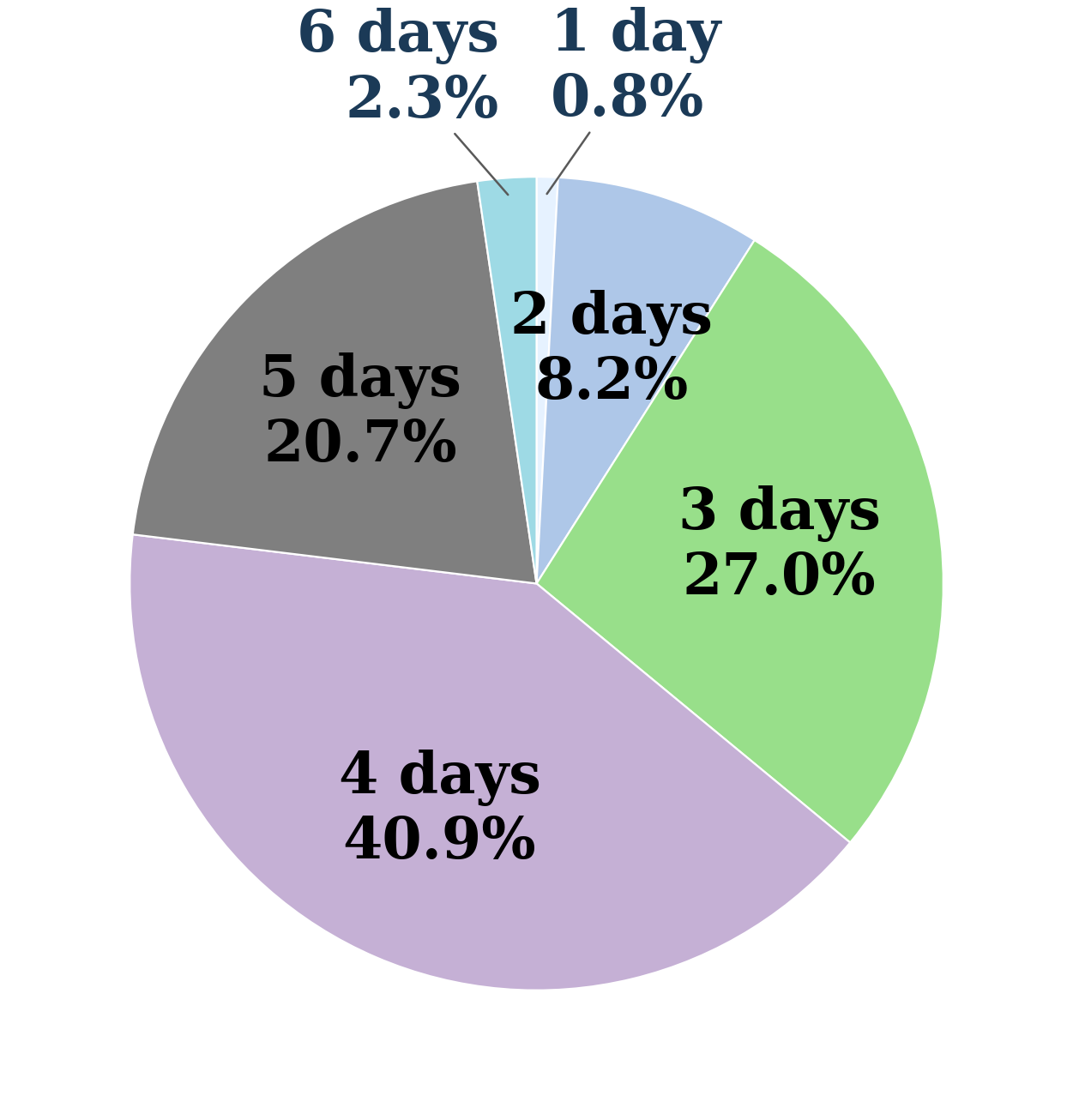}
        \caption{Dataset: Stieger2021 Subj:28 Sim: TanDM}
    \end{subfigure}
    \begin{subfigure}{0.24\textwidth}
        \centering
        \includegraphics[width=\linewidth]{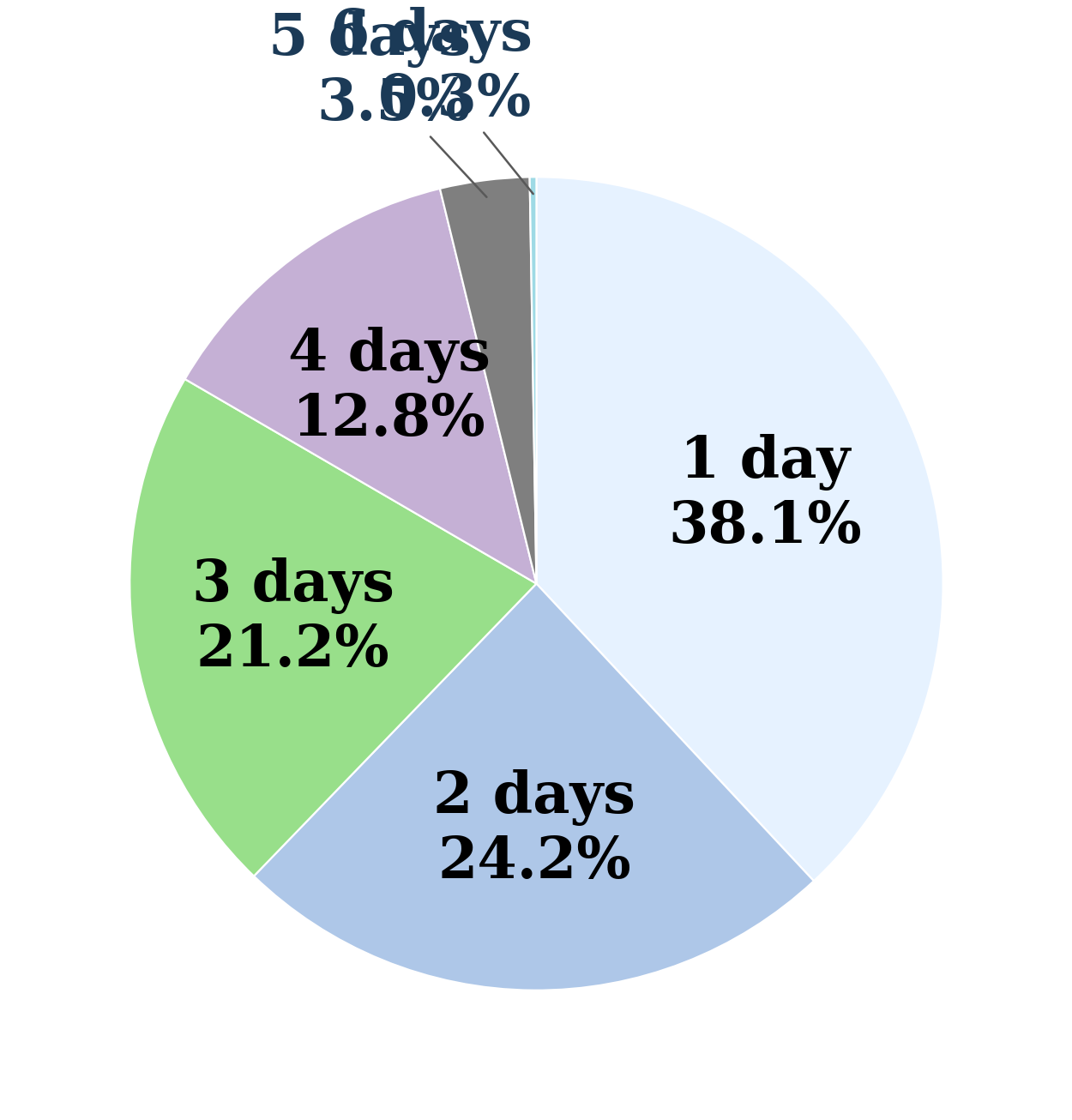}
        \caption{Dataset: Stieger2021 Subj:30 Sim: Cos}
    \end{subfigure}
    \begin{subfigure}{0.24\textwidth}
        \centering
        \includegraphics[width=\linewidth]{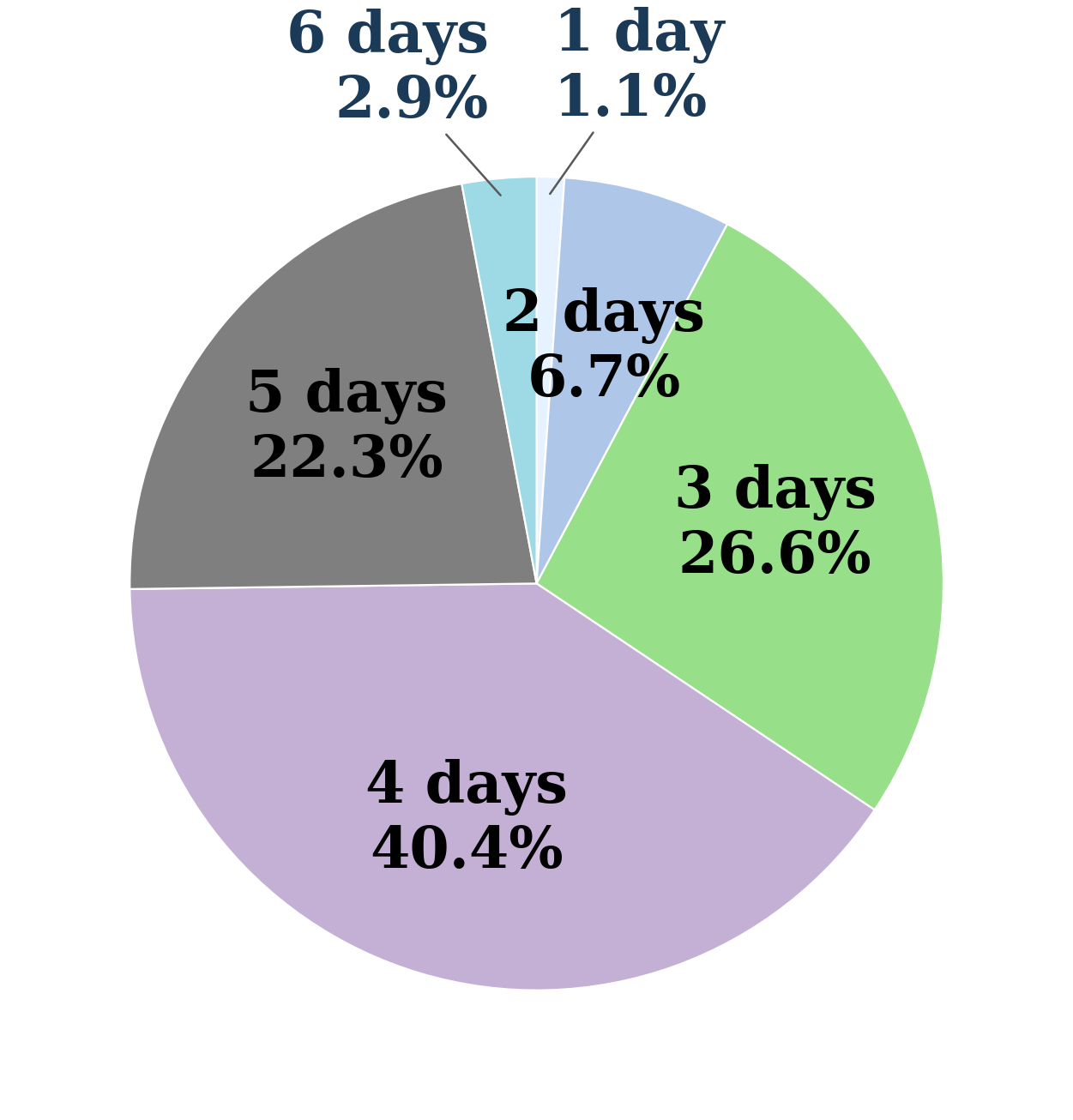}
        \caption{Dataset: Stieger2021 Subj:30 Sim: TanDM}
    \end{subfigure}
    \begin{subfigure}{0.24\textwidth}
        \centering
        \includegraphics[width=\linewidth]{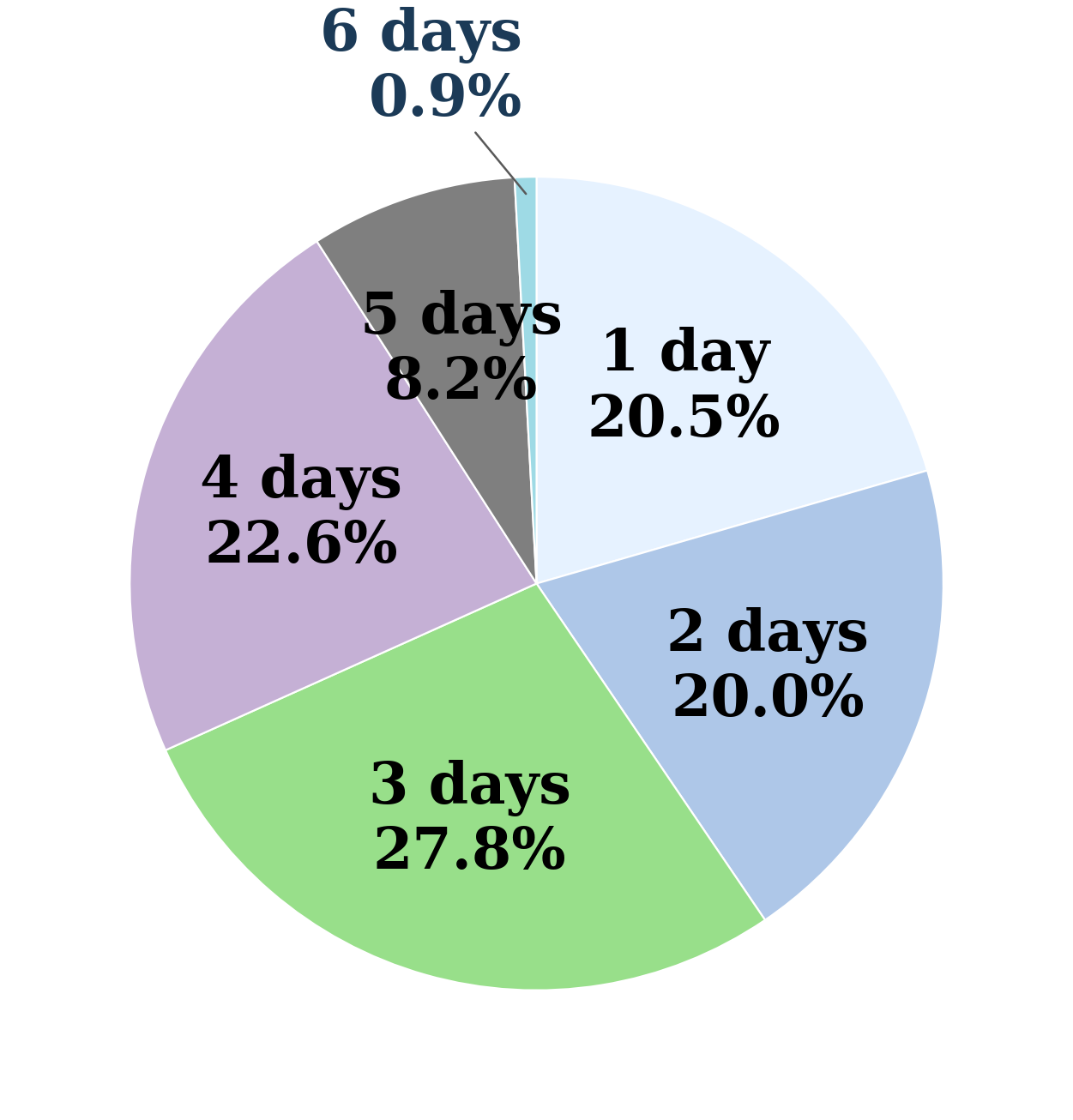}
        \caption{Dataset: Stieger2021 Subj:46 Sim: Cos}
    \end{subfigure}
    \begin{subfigure}{0.24\textwidth}
        \centering
        \includegraphics[width=\linewidth]{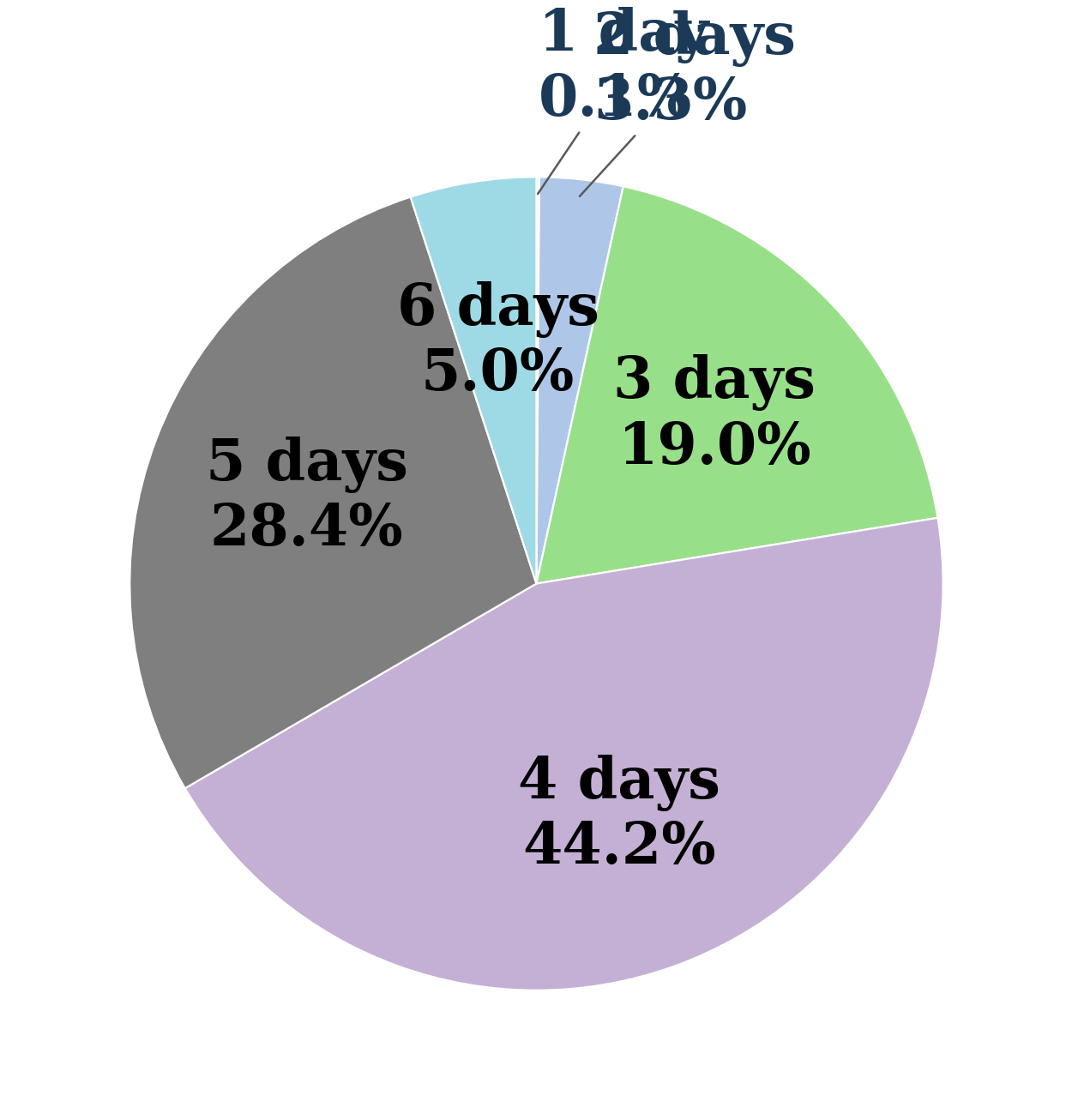}
        \caption{Dataset: Stieger2021 Subj:46 Sim: TanDM}
    \end{subfigure}
    \begin{subfigure}{0.24\textwidth}
        \centering
        \includegraphics[width=\linewidth]{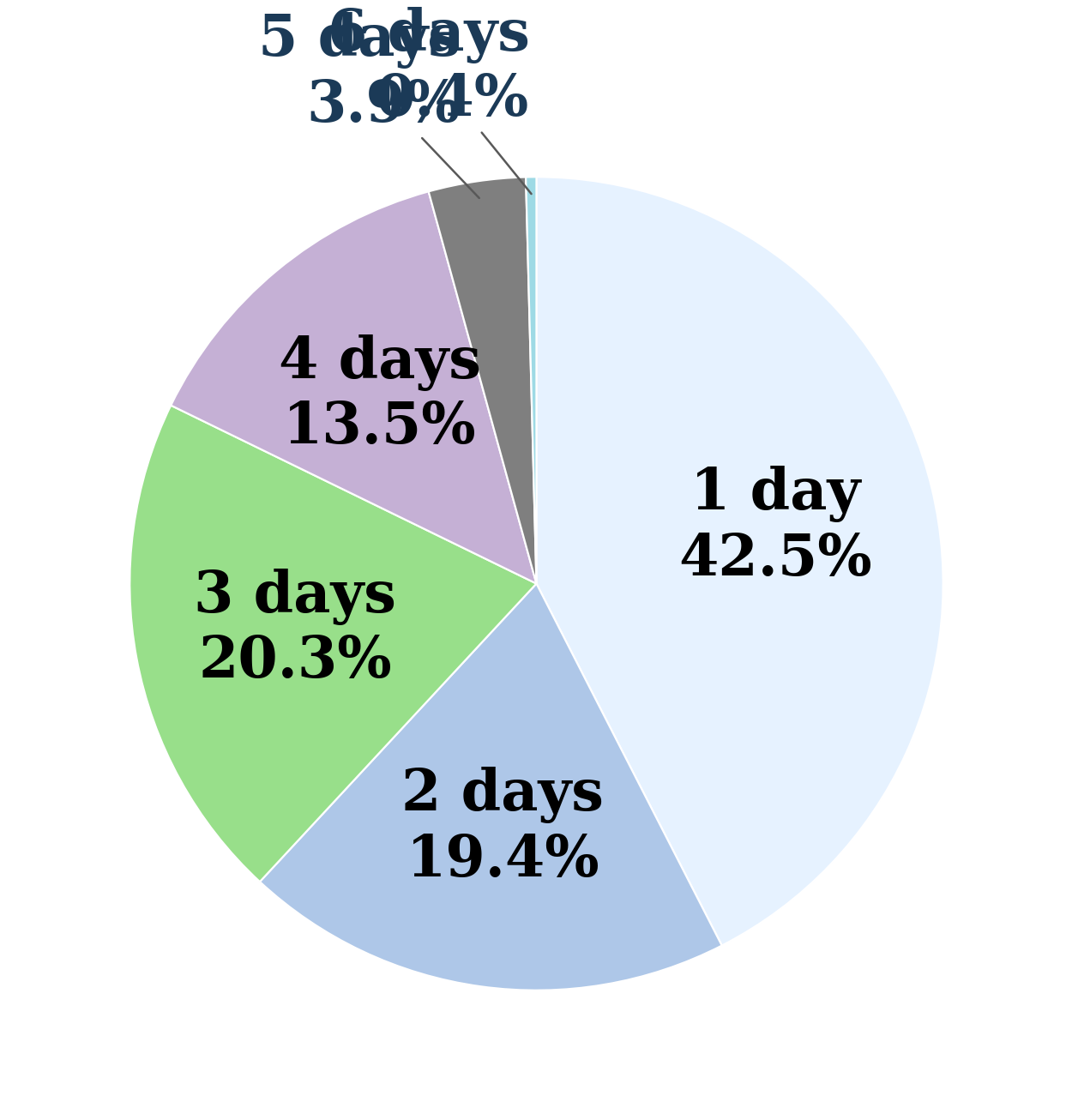}
        \caption{Dataset: Stieger2021 Subj:50 Sim: Cos}
    \end{subfigure}
    \begin{subfigure}{0.24\textwidth}
        \centering
        \includegraphics[width=\linewidth]{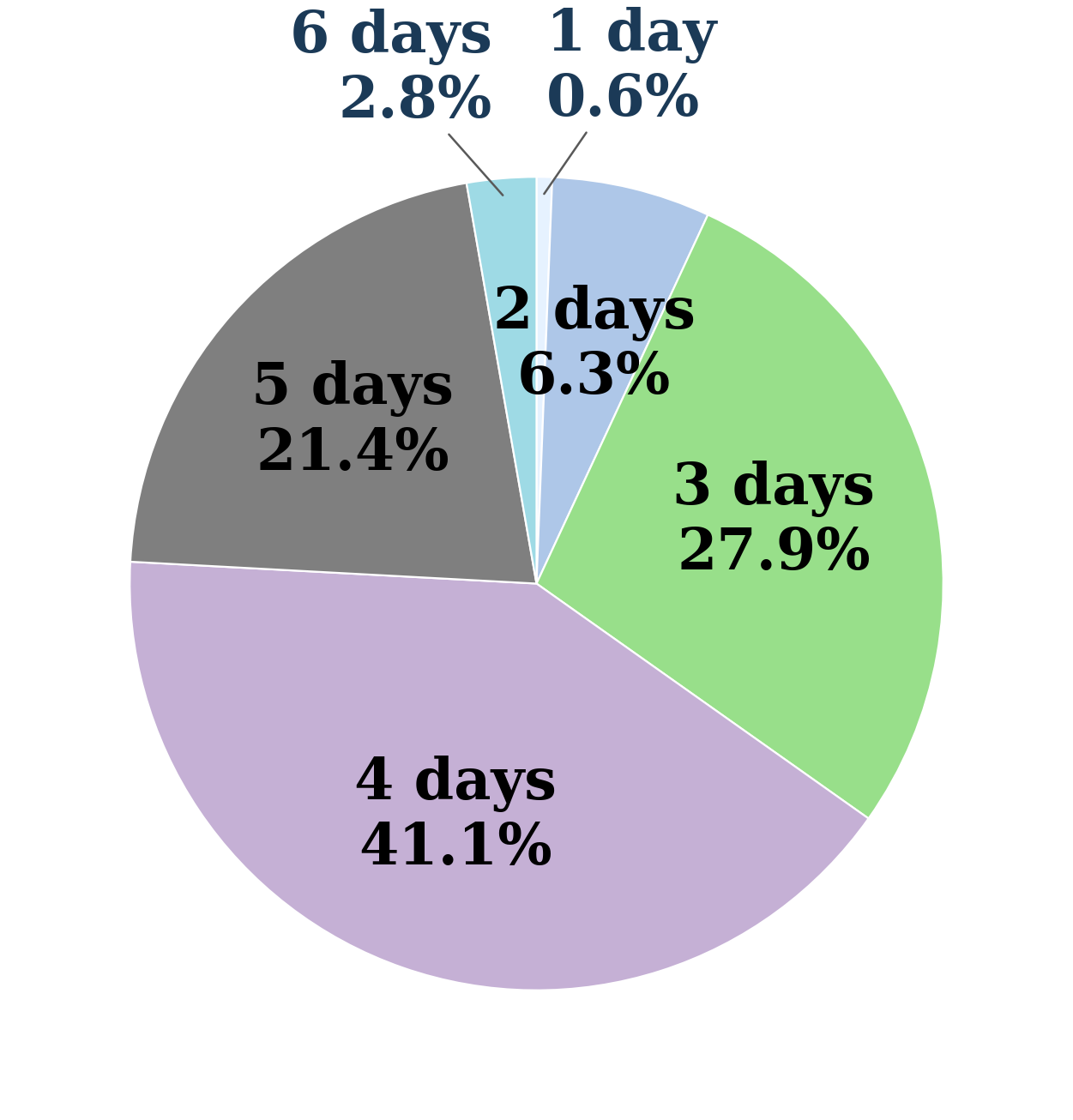}
        \caption{Dataset: Stieger2021 Subj:50 Sim: TanDM}
    \end{subfigure}    
    \caption{Percentage of distinct days per hyperedge in the covariance‑based hypergraph component}
    \label{fig:pie_chart_s}
\end{figure}

% \newpage
% \input{checklist.tex}

\end{document}